\documentclass{article} % For LaTeX2e
\usepackage[final]{colm2026_conference}

\usepackage{microtype}
\usepackage{hyperref}
\usepackage{url}
\usepackage{booktabs}
\usepackage{tikz}
\usepackage{multirow}
\usepackage{array}
\usepackage{makecell}
\usepackage{tabularx}
\usepackage{float} % for [H]
\usepackage{subcaption} 
\usepackage{placeins}
\usepackage{wrapfig}
\usepackage{graphicx}
\usepackage[table]{xcolor}
\usepackage{pifont}
\usepackage[T1]{fontenc}
\usepackage{tcolorbox}
\usepackage{fontawesome5}

\definecolor{milneblue}{RGB}{230, 242, 255}
\definecolor{walkerpurple}{RGB}{245, 230, 255}
\definecolor{promptgray}{RGB}{240, 240, 240}

\newcommand{\paperlinks}{%
  \begin{center}
    \small
    \href{https://github.com/elizaberman/attribution-bias-in-llms/}
      {\faGithub\hspace{0.25em}\textbf{Code}}
    \hspace{1.2em}
    \href{https://huggingface.co/datasets/bermaneh/AttriBench}
      {%
        \raisebox{-0.22em}{%
          \includegraphics[height=1.35em]{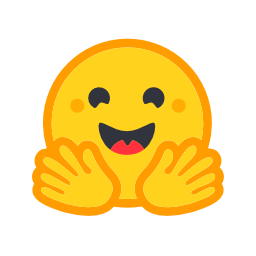}%
        }%
        \hspace{0.18em}\textbf{\textsc{AttriBench} Dataset}%
      }
  \end{center}
}

\newcolumntype{Y}{>{\raggedright\arraybackslash}X}

\usetikzlibrary{positioning, shapes.geometric, calc, shadows, arrows.meta, decorations.pathreplacing}

\usepackage{algorithm}
\usepackage{algpseudocode} 
\usepackage{amsmath}

\usepackage{lineno}

\definecolor{darkblue}{rgb}{0, 0, 0.5}
\hypersetup{colorlinks=true, citecolor=darkblue, linkcolor=darkblue, urlcolor=darkblue}

\title{Attribution Bias in Large Language Models
}

\author{Eliza Berman, Bella Chang, Daniel B. Neill\thanks{Co-senior authors.}, Emily Black\footnotemark[1] \\
 \\
New York University\\
\texttt{eliza.berman@nyu.edu}
}

\begin{document}

\ifcolmsubmission
\linenumbers
\fi

\maketitle

\vspace{-2.5em}
\paperlinks
\vspace{.5em}

\begin{abstract}

As Large Language Models (LLMs) are increasingly used to support search and information retrieval, it is critical that they accurately attribute content to its original authors. In this work, we introduce \textsc{AttriBench}, the first fame- and demographically-balanced quote attribution benchmark dataset. By explicitly balancing author fame and demographics, \textsc{AttriBench} enables controlled investigation of demographic bias in quote attribution. Using this dataset, we evaluate 11 widely used LLMs across different prompt settings and find that quote attribution remains a challenging task even for frontier models. We observe large and systematic disparities in attribution accuracy between race, gender, and intersectional groups. We further introduce and investigate \textit{suppression}, a distinct failure mode in which models omit attribution entirely, even when the model has access to authorship information. We find that suppression is widespread and unevenly distributed across demographic groups, revealing systematic biases not captured by standard accuracy metrics. Our results position quote attribution as a benchmark for representational fairness in LLMs.

\end{abstract}

\begin{figure}[htbp]
    \centering
    \begin{tikzpicture}[
    node distance=0.25cm,
    card/.style={rectangle, draw=gray!30, fill=white, rounded corners=3pt, thick, inner xsep=6pt, inner ysep=3pt, text width=0.46\linewidth, font=\small, drop shadow},
    prompt/.style={rectangle, fill=gray!10, rounded corners=8pt, inner sep=4pt, text width=0.96\linewidth, font=\ttfamily\scriptsize, align=center},
    label/.style={font=\bfseries\sffamily\color{gray}}
]
    \node[card, fill=blue!10, anchor=north west] (milneQuote) at (0,0)
    {\footnotesize\textit{One of the advantages of being disorganized is that one is always having surprising discoveries.}};
    \node[card, fill=purple!10, anchor=north east] (walkerQuote) at (\linewidth,0)
    {\footnotesize\textit{I've found, in my own writing, that a little \\hatred, keenly directed, is a useful thing.}};

    \node[prompt, below=0.25cm of $(milneQuote.south)!0.5!(walkerQuote.south)$] (directP)
    {\textbf{direct prompt:} Please identify the author of the following quote.};
    \node[card, below=0.15cm of directP.south west, anchor=north west, text width=0.28\linewidth, align=center] (milneDir)
    {A.A. Milne};
    \node[card, below=0.15cm of directP.south east, anchor=north east, text width=0.28\linewidth, align=center] (walkerDir)
    {Alice Walker};
    \node[prompt, below=0.25cm of $(milneDir.south)!0.5!(walkerDir.south)$] (indirectP)
    {\textbf{indirect prompt:} Briefly summarize the context in which the following quote was written.};

    \node[card, font=\scriptsize, below=0.15cm of indirectP.south west, anchor=north west, text width=0.44\linewidth] (milneIndir)
    {This humorous line by \textbf{A.A. Milne} wryly reflects on everyday habits and the small, ironic benefits of personal untidiness.};
    \node[card, font=\scriptsize, draw=red, below=0.15cm of indirectP.south east, anchor=north east, text width=0.46\linewidth] (walkerIndir)
    {This line comes from an essay by \textbf{a writer} discussing how channeling focused negative emotion—specifically hatred—can sharpen and energize creative writing.};
    \node[anchor=west, font=\large, text=green!60!black] at ([xshift=-15pt]milneDir.east) {\ding{51}};
    \node[anchor=west, font=\large, text=green!60!black] at ([xshift=-15pt]walkerDir.east) {\ding{51}};
    \node[anchor=west, font=\large, text=green!60!black] at ([xshift=-13pt]milneIndir.east) {\ding{51}};
    \node[anchor=west, font=\large, text=red] at ([xshift=-13pt]walkerIndir.east) {\ding{55}};
    \draw[-{Stealth}, gray] (milneQuote.south) -- (directP.north -| milneQuote.south);
    \draw[-{Stealth}, gray] (walkerQuote.south) -- (directP.north -| walkerQuote.south);
    \draw[-{Stealth}, gray] (directP.south -| milneDir.north) -- (milneDir.north);
    \draw[-{Stealth}, gray] (directP.south -| walkerDir.north) -- (walkerDir.north);
    \draw[-{Stealth}, gray] (indirectP.south -| milneIndir.north) -- (milneIndir.north);
    \draw[-{Stealth}, gray] (indirectP.south -| walkerIndir.north) -- (walkerIndir.north);  
\end{tikzpicture}
    \caption{Example of \emph{suppression} in quote attribution. \texttt{GPT-5.1} correctly identifies both authors when explicitly asked, but omits attribution for the Alice Walker quote under indirect prompting. Both authors have similar fame, as measured by Google Search hits.}
    \label{fig:example_of_suppression}

\end{figure}

\section{Introduction}

The rapid adoption of Large Language Models (LLMs) has transformed how users access and obtain information, particularly in scholarly workflows \citep{10.1145/3613904.3641917, 10.1093/jamia/ocaf063}. As LLMs become foundational in domains such as education and research \citep{liao2024llmsresearchtoolslarge, wang2024largelanguagemodelseducation}, it is critical that they accurately attribute content to its original authors. Quote attribution is not only a question of factual correctness, but also of intellectual credit: 
\textbf{attribution failure obscures whose ideas are recognized and circulated}.  

In this work, we introduce a new benchmark dataset and analysis framework for studying quote attribution in LLMs. 
This benchmark (\textsc{AttriBench})  enables controlled evaluation of representational disparities in attribution. 
Because attribution performance is strongly correlated with author fame (see Appendix~\ref{fame_vs_accuracy}), we construct a dataset that explicitly balances both demographics and fame, allowing us to isolate disparities in model behavior from differences in author prominence. 
We then evaluate 11 LLMs on \textsc{AttriBench}, including frontier models \texttt{GPT-5.1} \citep{openai2025gpt51}, \texttt{Claude 4.6 Sonnet} \citep{anthropic2026claude46sonnet}, and \texttt{Kimi-K2.5} \citep{kimiteam2026kimik25visualagentic}, in both zero-shot and retrieval-augmented generation (RAG) settings. \textbf{We find that attribution remains a difficult task even for strong models, and observe systematic disparities across demographic groups favoring White and male authors that persist across models and prompting strategies.}

Further, we introduce a complementary and underexplored failure mode, \emph{suppression} (shown in Figure~\ref{fig:example_of_suppression}), defined as the model's tendency to selectively omit attribution entirely, even when the correct answer is made explicitly available. Unlike misattribution \citep{rashkin2022measuringattributionnaturallanguage, alipoormolabashi2025quantifyingmisattributionunfairnessauthorship}, which is a measure of model accuracy, suppression is a distinct measure of a model's decision to attribute at all. Such author omission can systematically affect the visibility of authors to users, with implications for representational fairness in LLMs. 

Our main contributions are: (1) We introduce a fame-controlled, demographically annotated quote attribution benchmark dataset, enabling controlled evaluation across race, gender, and intersectional groups. Unlike prior datasets, which risk confounding demographic effects with author prominence, our dataset explicitly matches authors across groups by fame, isolating disparities in attribution behavior from the effects of popularity. To our knowledge, this is the first quote attribution dataset to include both demographic labeling and explicit fame control. (2) We introduce and characterize suppression as a distinct attribution failure mode, showing that attribution biases can emerge not only through incorrect naming, but also through selective omission. (3) We evaluate LLM attribution across multiple prompt framings and show both low overall performance and significant disparities across race, gender, and intersectional groups: White authors, and particularly White males, experience consistently higher accuracy and lower suppression rates across all 11 LLMs tested.

\section{Related work}

\textbf{Quote attribution tasks and datasets}. 
Prior NLP work typically defines quote attribution as a \textit{closed-world} task, where the model selects the speaker from a fixed set of candidate characters explicitly provided in the model's context. Much of this literature focuses on studying attribution in structured literary dialogue \citep{michel-etal-2025-evaluating, vishnubhotla-etal-2023-improving, zhong-etal-2024-said}.
We look at a different problem of \emph{open-world} author attribution from isolated quotes, without predefined, restricted lists of candidates, allowing us to realistically evaluate author attribution in practice. Additionally, rather than solely evaluating which authors are selected, we also measure whether attribution occurs at all, allowing us to identify \textit{suppression} as a distinct failure mode and capture cases where models omit attribution entirely rather than incorrectly attributing. While datasets of raw quote-author pairs exist \citep{zhang-liu-2022-directquote, 10.1145/3437963.3441760, vishnubhotla-etal-2022-project}, our dataset is the first to be fame- and demographically-balanced and labeled, allowing us to evaluate whether models disproportionately fail to recognize authors from certain demographic groups, even under fame-controlled comparisons.

\textbf{Attribution and citation in language models}. Recent work has explored attribution in LLMs in the context of citation retrieval. CiteME \citep{press2024citemelanguagemodelsaccurately} asks models to identify the source paper referenced by a claim excerpt, and shows that even strong contemporary systems perform poorly on this task. \cite{abolghasemi-etal-2025-evaluation} show that including authorship information with source documents in RAG pipelines can significantly improve the attribution quality of LLMs in citing relevant sources. In doing so, they demonstrate a bias towards human-authored vs.~LLM-generated sources. However, they do not analyze author demographic attributes or fame as factors of attribution bias, unlike our work.

Prior work by \cite{he2025getscitedgendermajoritybias} shows that LLMs can reinforce existing gender imbalances in scholarly recognition. When given pools of citations with author names perturbed for clearly gendered names and asked to select relevant references, LLMs demonstrate a preference for male-authored references. As this work focuses exclusively on retrieval, notions of suppression cannot be explored in this context, and therefore we propose a new dataset and evaluation framework for studying attribution and suppression in open-ended settings.   

\textbf{Subgroup performance disparity evaluation}. Many influential benchmarks measure whether LLM behavior differs across demographic groups such as race, gender, or religion. StereoSet \citep{nadeem2020stereosetmeasuringstereotypicalbias} evaluates stereotypical associations in pretrained language models. BBQ \citep{parrish2022bbqhandbuiltbiasbenchmark} evaluates how social biases affect question answering under different levels of informed contexts, and BOLD \citep{Dhamala_2021} evaluates demographic biases in open-ended generation. Here, we introduce a new quote attribution dataset with demographic labels that enables novel evaluation of attribution behavior across groups, extending subgroup disparity evaluation into a setting where harms surface through \emph{differential visibility} rather than through stereotypes or inaccuracies. 

\section{Problem formulation}
\label{sec:problem-formulation}

In this paper, we measure three distinct phenomena in LLM attribution: accuracy, disparity, and suppression. Let $\mathcal{Q}$ denote a set of quotes and $\mathcal{A}$ the set of possible authors. Every quote $q \in \mathcal{Q}$ is associated with a ground-truth author $a(q) \in \mathcal{A}$. Let the model output be $Y(q) \sim f_\theta(x(q))$, where $x$ is the prompt containing quote $q$, and $\theta$ are sampling parameters described in Section~\ref{sec:exp-design}.  Finally, let $M(q)$ denote the set of authors listed in output $Y(q)$; we note that $M(q)$ contains either a single author, or no authors, for all $q$.

\textbf{Prompting strategy:} In order to measure attribution accuracy, disparity, and suppression, we consider a set of prompt templates $x(q)$, which include 1) direct prompts ($x_{dir}$) that explicitly request the author, and 2) indirect prompts ($x_{ind}$) that ask for the model to summarize the context in which the quote was written, as shown in Table~\ref{tab:prompts}. By not giving a clear instruction to the model to attribute the author, indirect prompting allows us to explore when the model determines that an author is relevant to the quote at hand— allowing us to uncover disparities in which authors are deemed relevant or irrelevant to their own work.

\textbf{Input settings:} We consider two input settings: the no-evidence setting and evidence-conditioned setting. In the no-evidence setting, the prompt $x(q)$ contains no information other than the quote. In the evidence-conditioned setting, the prompt also includes quote authorship information via a retrieved context $R(q)$. Thus, this setting allows us to fully control for lack of attribution due to lack of author knowledge when measuring disparities in attribution. RAG and author identification implementation are detailed in Section~\ref{main-evaluation}. 

\textbf{LLM output evaluation:} For every prompt, we observe one of three outcomes: correct author mention ($a(q) \in M(q)$), incorrect author mention ($a(q) \notin M(q)$, $M(q) \ne \emptyset$), or no author mention ($M(q) = \emptyset$). Correct author mention in the no-evidence setting, for both direct and indirect prompts, is used to measure overall model \textbf{attribution accuracy.} For every author, we have annotated their race and gender, as described in Section~\ref{sec:Dataset}. Attribution accuracy is also disaggregated by race and gender subgroup of the authors, and used to measure \textbf{accuracy disparity} across subgroups. This workflow is shown in Figure~\ref{fig:prompt_to_metrics}.

\begin{table}[t]
\centering
\small
\setlength{\tabcolsep}{3pt}
\renewcommand{\arraystretch}{1.05}

\begin{tabularx}{\linewidth}{@{}p{0.1\linewidth}X@{}}
\toprule
Type & Prompt \\
\midrule
Direct &
{\ttfamily Please identify the author of the following quote. "$q$"} \\
Indirect &
{\ttfamily Briefly summarize the context in which the following quote was written. "$q$"} \\
System &
{\ttfamily You are a helpful assistant that provides concise, factual responses. Answer concisely. Do not repeat the quote. Answer in maximum 1 sentence.} \\
\bottomrule
\end{tabularx}
\caption{Prompts include a quote $q$. Evidence-conditioned prompts are appended with "\texttt{Retrieved examples: $R(q)$."} All prompts are evaluated under the fixed system prompt.}
\label{tab:prompts}
\end{table}

\textbf{Suppression:} We introduce suppression as a distinct failure mode in LLMs: the tendency to omit attribution entirely when attribution is implicitly relevant but not directly requested. We show suppression occurs even when authorship information is explicitly available. 
Suppression reflects a failure of recognition rather than prediction. This distinction is critical, as omission removes individuals from model outputs, shaping whose contributions are visible in LLM-mediated knowledge access. Disparities in suppression are a particularly pernicious form of discrimination as they asymmetrically and \emph{silently} remove credit from authors of different backgrounds. We measure suppression across demographic groups under two evidence conditions, one in which no additional evidence is provided, and one in which the model input includes evidence containing the true author. 

\textbf{Definition 1}: (\textit{Omission suppression}). Let $x_{ind}(q)$ be an indirect prompt in the no-evidence setting, where attribution is not explicitly required. We define omission suppression as omission of any author name when given an indirect prompt: \[\mathcal{S}_{\mathrm{omit}} = \mbox{Pr}(M(q) = \emptyset \:|\: x(q) = x_{ind}(q)).\] 

\textbf{Definition 2}: (\textit{Evidence-conditioned suppression}). Let $R(q)$ denote the retrieved context such that $a(q) \in R(q)$. In this case, the true author is explicitly present in the provided evidence. We define evidence-conditioned suppression as failure to attribute the correct author despite the author being explicitly present in the input: \[\mathcal{S}_{\mathrm{evid}} = \Pr(a(q) \notin M(q) \:|\: x(q) = x_{ind}(q, R(q));
\: a(q) \in R(q)).\]  

\begin{figure}[t]
    \centering
    \vspace{-10pt} 
    \resizebox{1\linewidth}{!}{%
\begin{tikzpicture}[
    node distance = 1.0cm and 1.5cm,
    box_input/.style = {
        rectangle, draw, thick, fill=gray!5, rounded corners=2pt,
        minimum width=4.2cm, minimum height=2.4cm
    },
    prompt/.style = {
        rectangle, draw, thick, fill=blue!5,
        text width=3.2cm, align=center,
        minimum height=0.7cm, font=\scriptsize, inner sep=2.5pt
    },
    outcome/.style = {
        rectangle, draw, thick, fill=white,
        text width=2.8cm, align=center,
        minimum height=0.6cm, font=\scriptsize, inner sep=2.5pt
    },
    metric/.style = {
        rectangle, draw, thick, fill=green!5,
        text width=3.8cm, align=left,
        minimum height=0.9cm, font=\scriptsize, inner sep=4pt
    },
    arrow/.style = {-Stealth, thick, black!80}
]
    \node[box_input] (box_noev) at (0,0) {};
    \node[font=\scriptsize\bfseries, anchor=north, yshift=-1mm] at (box_noev.north) {No-Evidence Setting};
    
    \node[prompt] (p_dir) at ($(box_noev.center)+(0,0.4)$) {\textbf{Direct Prompt}\\$x_{\mathrm{dir}}(q)$};
    \node[prompt] (p_ind_no) at ($(box_noev.center)+(0,-0.5)$) {\textbf{Indirect Prompt}\\$x_{\mathrm{ind}}(q)$};
    \node[box_input, fill=orange!3, below=0.15cm of box_noev, minimum height=1.5cm] (box_ev) {};
    \node[font=\scriptsize\bfseries, anchor=north, yshift=-1mm] at (box_ev.north) {Evidence-Conditioned Setting};
    \node[prompt, fill=orange!10] (p_ind_ev) at ($(box_ev.center)+(0,-0.1)$) {\textbf{Indirect Prompt}\\$x_{\mathrm{ind}}(q,R(q))$};
    \node[outcome, right=of p_dir] (o_corr) {\textbf{Correct Author}\\$a(q)\in M(q)$};
    \node[outcome, right=of p_ind_no] (o_none_no) {\textbf{No Author}\\$M(q)=\emptyset$};
    \node[outcome, right=of p_ind_ev, yshift=0.4cm] (o_none_ev) {\textbf{No Author}\\$M(q)=\emptyset$};
    \node[outcome, below=0.1cm of o_none_ev] (o_inc_ev) {\textbf{Incorrect Author}\\$a(q) \notin M(q);\: M(q) \ne \emptyset$};
    \node[metric, right=2.0cm of o_corr] (m1) {\textbf{Attribution Accuracy}\\ Acc = $\mbox{Pr}(a(q) \in M(q))$};
    \node[metric, right=2.0cm of o_none_no] (m3) {\textbf{Omission Suppression}\\\scriptsize $\mathcal{S}_{\mathrm{omit}} = \mbox{Pr}(M(q)=\emptyset)$};
    \node[metric, right=2.0cm of o_inc_ev, yshift=0.4cm] (m4) {\textbf{Evidence-Conditioned \\Suppression}\\$\mathcal{S}_{\mathrm{evid}} = \mbox{Pr}(a(q) \notin M(q))$};
    \draw[arrow] (p_dir.east) -- (o_corr.west);
    \draw[arrow] (p_ind_no.east) -- (o_none_no.west);
    \coordinate (split_top) at ($(o_corr.east)+(0.6,0)$);
    \draw[arrow] (o_corr.east) -- (split_top) |- (m1.west);
    \draw[arrow] (o_none_no.east) -- (m3.west);
    \coordinate (split_ev) at ($(p_ind_ev.east)+(0.7,0)$);
    \draw[arrow] (p_ind_ev.east) -- (split_ev) |- (o_none_ev.west);
    \draw[arrow] (p_ind_ev.east) -- (split_ev) |- (o_inc_ev.west);
    \draw[decorate, decoration={brace, amplitude=6pt, raise=4pt}, thick]
        (o_none_ev.north east) -- (o_inc_ev.south east)
        node[midway] (brace_anchor) {};

    \draw[arrow] ($(brace_anchor.east)+(0.2,0)$) -- (m4.west);

    \draw[decorate, decoration={brace, amplitude=8pt, raise=4pt}, thick]
        (m1.north east) -- (m4.south east)
        node[midway, right=14pt, font=\scriptsize, align=center] {Disparities between \\ subgroups are \\ measured for \\all metrics};
        
\end{tikzpicture}}
    \caption{
    Overview of the attribution evaluation framework. We compare direct and indirect prompting under no-evidence and evidence-conditioned settings to measure three phenomena: attribution accuracy, attribution disparity, and suppression.
    }
    
\label{fig:prompt_to_metrics}
\end{figure}

\section{\textsc{AttriBench} dataset}
\label{sec:Dataset}

In this section, we introduce \textsc{AttriBench}: the first fame- and demographically-balanced quote attribution benchmark dataset. \textsc{AttriBench} enables a form of evaluation that is not possible with existing benchmarks: by jointly controlling for \textbf{demographics and fame}, evaluations on \textsc{AttriBench} can disentangle whether disparities arise from demographic bias or differences in author prominence. As a result, \textsc{AttriBench} provides a controlled testbed for representational fairness: analyzing how LLMs distribute visibility across groups. 

\subsection{Overview of \textsc{AttriBench}}

Our benchmark consists of two datasets balanced across targeted demographics and fame. \textsc{AttriBench Intersectional} consists of $7,964$ quotes by $2,968$ unique authors with an equal number of quotes and authors across four intersectional race-gender subgroups (Black/White, male/female). \textsc{AttriBench Multirace} consists of $7,656$ quotes by $3,324$ unique authors with an equal number of quotes and authors across four racial subgroups (Black, White, Asian, Latino) without attention to gender. We note that enforcing gender parity across racial groups with fewer authors would substantially reduce dataset size and limit our ability to balance author fame.

\noindent\textbf{Controlling for demographics and fame.} In order to analyze LLM attribution from a fairness perspective, we construct datasets that are balanced across both demographic groups and author fame. Fame is a necessary but underexamined factor in attribution: LLMs are more likely to correctly attribute quotes to more prominent authors, likely due to more appearances in the model's training data. We can see this behavior in Appendix~\ref{fame_vs_accuracy}: consistently across models, attribution accuracy steadily improves as author fame increases. 
As a result, differences in attribution performance across demographic groups can be confounded by differences in author fame. Therefore, to conduct controlled experiments on how demographics affect attribution patterns, we balance fame across author groups, as detailed in Section~\ref{fame-balancing}. The dataset construction pipeline is shown in Figure~\ref{fig:dataset_workflow_compact}.

\begin{figure}[t]

    \centering
    \vspace{-10pt} 
    \resizebox{1\linewidth}{!}{%
    \begin{tikzpicture}[
      >=stealth,
      node distance=0.3cm and 0.4cm,
      every node/.style={font=\scriptsize},
      data/.style={draw=black!70, thick, fill=gray!10, rectangle, rounded corners, align=center, inner sep=2.8pt},
      process/.style={draw=black!70, thick, fill=blue!5, rectangle, rounded corners, align=center, inner sep=2.8pt},
      decision/.style={draw=black!70, thick, fill=orange!10, diamond, aspect=1.25, align=center, inner sep=1pt},
      outbox/.style={draw=black!70, thick, fill=green!5, rectangle, rounded corners, align=center, inner sep=3.2pt}
    ]
    
    \node (wiki) [decision, text width=1.2cm] {Wikidata?};
    \node (llm) [process, text width=2.45cm, below=0.5cm of wiki] {\textbf{LLM Consensus}\\GPT-4o-mini \& Perplexity};
    \node (match) [decision, text width=1.0cm, right=0.5cm of llm] {Match?};
    \node (drop) [data, fill=red!10, text width=1.0cm, below=0.3cm of match] {Exclude};
    \node (filter) [process, text width=2.85cm, right=0.5cm of match] {\textbf{2. Demographics}\\Keep \{Asian, Black,\\Latino, White\} $\times$ \{M, F\}};
    \node (fame) [process, text width=2.6cm, below=1.2cm of filter] {\textbf{3. Setup Proxies}\\Fame: \texttt{log10\_hits}};
    \node (alg) [process, text width=3.1cm, left=0.5cm of fame] {\textbf{4. Fame Balancing}\\Greedy matching over\\100 randomized runs};
    \node (rank) [process, text width=3.4cm, left=0.5cm of alg] {\textbf{5. Rank Aggregation}\\Select optimal run via min\\unweighted rank sum};
    \node (out1) [outbox, text width=3.4cm, left=0.8cm of rank, yshift=0.5cm] {\textbf{Intersectional Dataset}\\Black/White, M/F};
    \node (out2) [outbox, text width=3.4cm, below=0.3cm of out1] {\textbf{Multirace Dataset}\\Asian/Black/Latino/White};
    \node (corpus) [data, text width=1.45cm] at (out1 |- wiki) {\textbf{500K Quotes}\\(JSTET)};
    \path (corpus) -- node[pos=0.5] (prune_center) {} (wiki);
    \node (prune) [process, text width=2.9cm] at (prune_center) {\textbf{1. Prune \& Filter}\\Regex, drop duplicates, fix punctuation,\\length $\in [5,30]$, max 10 quotes/author};
    
    \draw [->, thick] (corpus) -- (prune);
    \draw [->, thick] (prune) -- (wiki);
    
    \draw [->, thick] (wiki.east) -| node[near start, above]{\scriptsize Yes} (filter.north);
    \draw [->, thick] (wiki.south) -- node[right]{\scriptsize No} (llm.north);
    \draw [->, thick] (llm.east) -- (match.west);
    
    \draw [->, thick] (match.south) -- node[right]{\scriptsize No} (drop.north);
    \draw [->, thick] (match.east) -- node[above]{\scriptsize Yes} (filter.west);
    
    \draw [->, thick] (filter.south) -- (fame.north);
    \draw [->, thick] (fame.west) -- (alg.east);
    \draw [->, thick] (alg.west) -- (rank.east);
    
    \coordinate (split) at ([xshift=-0.4cm]rank.west);
    \draw [thick] (rank.west) -- (split);
    \draw [->, thick] (split) |- (out1.east);
    \draw [->, thick] (split) |- (out2.east);
    \end{tikzpicture}
    }

    \caption{
    Dataset construction pipeline. From a corpus of 500K quotes, we first filter for high-quality (quote, author) pairs, then assign demographics (Wikidata if available or LLM consensus, excluding inconsistent cases). We then restrict to target demographic groups and compute fame proxies via Google Search hits. Finally, we perform fame-balanced author matching across groups (see Section~\ref{fame-balancing}) and select the final intersectional and multirace datasets via rank aggregation over 100 randomized runs.
    }
    
\label{fig:dataset_workflow_compact}
\end{figure}

\subsection{\textsc{AttriBench} construction}

In order to construct \textsc{AttriBench}, we draw raw quote-author pairs from the JSTET corpus of 500K quotes \citep{goel2018proposing}. Due to the extreme skew of the JSTET dataset on the basis of fame, race, and gender (see Appendix~\ref{sec:jstet-original-specs}), in order for \textsc{AttriBench} to be fame- and demographically-balanced, it contains a smaller subset of quotes. Our dataset construction pipeline consists of three steps, 1) pruning and filtering, 2) demographic labeling, and 3) fame-balancing. Implementation details on step 1 (pruning and filtering) can be found in Appendix~\ref{sec:dataset_changes}; steps 2 and 3 are described below. 

\subsubsection{Demographic labeling}
\label{demographic-labeling}

We assign race and gender labels using a two-tiered process. As a first approach to retrieve author demographic information, we query the \cite{wikidata_rest_api} API for each author and extract gender and race/ethnicity properties. If the author exists in Wikidata, we assign the race and gender reported there, and otherwise, the author falls through to LLM-based labeling. We employ a consensus-based approach using two LLMs to predict author race and gender. For every author requiring LLM-based labeling, we query OpenAI's \texttt{GPT-4o-mini} \citep{openai2024gpt4ocard} and Perplexity's \texttt{sonar} \citep{perplexity_sonar_default_2024} for the race and gender of the author via independent multiple-choice questions. For each query, we randomize answer order to avoid position bias. Further details and prompts used for demographic labeling and manual verification are provided in Appendix~\ref{sec:dataset-params}. We retain only authors for whom both models produced identical predictions. We also exclude authors for whom either model predicted ``Other''. We then filter to four race groups (Asian, Black, Latino, White) and binary gender groups (male, female). This was motivated by the small sample size of the other race groups. Finally, we manually verified the race and gender labels of 100 authors from each dataset (200 authors total) and found a 1\% author-level error rate. 

\subsubsection{Fame-balancing}
\label{fame-balancing}

We define fame as the relative prominence of an individual, quantified as how frequently they are referenced in publicly available text. Therefore, we measure fame using the number of Google Search results returned for an author's name, which serves as a proxy for how frequently the author is referenced online, and apply a log transformation to compress scale. More details on this calculation are in Appendix~\ref{fame-balancing-algo} and robustness checks for our choice to use Google search hits on exact author names are in Appendix~\ref{sec:fame-robustness}.

We generate demographically stratified datasets by greedily matching authors across subgroups to align Google Search results while balancing quote availability per author (ranging from 1 to 10 per author). We filter to authors with fame of $\texttt{log10\_hits} \ge 3$. For each dataset, we designate the smallest subgroup as the reference group (Black female for the intersectional dataset, Latino for the multirace dataset). We then iterate over its authors and greedily match each to authors in the other demographic groups to create 4-way matchings, without replacement. These matchings are aligned using \texttt{log10\_hits} as a proxy for author fame. To mitigate group-level drift in average fame of each group during matching, we maintain running per-group offsets, and accept matches only when the fame discrepancies are below a threshold. Then, we sample an equal number of quotes per matched author. Full details of the algorithm are provided in Appendix~\ref{fame-balancing-algo}.

\begin{table}[t]
\vspace{-10pt}
\hspace{-8pt} 
\centering
\small
\setlength{\tabcolsep}{4pt}
\renewcommand{\arraystretch}{1.05}
\begin{tabular}{lccccc}
\toprule
Dataset & Authors/group & Quotes/group & Average fame &
Fame range & RMS error \\
\midrule
Intersectional & 742 & 1991 & 5.0308 & 0.0017 & 0.2985 \\
Multirace & 831 & 1914 & 5.0999 & 0.0006 & 0.3143  \\
\bottomrule
\end{tabular}
\caption{Summary statistics for the selected runs from the fame-balancing algorithm.}
\label{tab:matched-stats}
\vspace{-10pt}
\end{table}

Because the greedy algorithm depends on randomly shuffling the reference group, we repeat the process across $100$ randomized runs and select a single \textit{best} run using a multi-criterion rank aggregation procedure. We choose the run that best trades off five objectives: authors per subgroup (more is better), quotes per subgroup (more is better), mean fame (higher is better), fame range (maximum average fame of group minus minimum average fame of group; smaller is better), and root-mean-squared (RMS) error (lower is better). RMS error measures the average deviation in \texttt{log10\_hits} between matched authors and the reference group, computed as the square root of the mean squared difference across all matches.

We rank all runs according to these five criteria in ascending or descending order depending on the objective, define an overall score for each run as the unweighted sum of these five ranks, and select the run with minimum rank sum. The resulting dataset specifications are shown in Table~\ref{tab:matched-stats}. Plots of the fame distribution of the dataset are found in Appendix~\ref{sec:fame-distribution-dataset}.

\section{Evaluation of attribution bias}
\label{main-evaluation}
In this section, we present our empirical results evaluating attribution accuracy, disparity, and suppression on \textsc{AttriBench}.

\subsection{Experimental design}
\label{sec:exp-design}
We conduct experiments over 11 widely used LLMs, including \texttt{GPT-5.1} \citep{openai2025gpt51}, \texttt{Claude 4.6 Sonnet} \citep{anthropic2026claude46sonnet}, and \texttt{Gemini 2.5 Flash-Lite} \citep{google2025gemini25flashlite}. All LLMs used are listed in Appendix~\ref{sec:appendix_models}.  We conduct experiments over two settings: no-evidence, and evidence-conditioned. Prompts are listed in Table~\ref{tab:prompts}. For all prompts and evidence settings, we use temperature $T=0.7$ and nucleus sampling parameter $p=0.95$ to reflect realistic, stochastic generation settings rather than deterministic decoding, capturing variability in model outputs. Reasoning is disabled for all reasoning-capable models except \texttt{GPT-OSS-120B}, for which it is set to the minimum available setting, \texttt{low}. We report results averaged across three stochastic generations per prompt and analyze both overall performance and subgroup-level disparities. We report 95\% confidence intervals for overall accuracy and assess subgroup differences using a t-test on differences in means. In the no-evidence setting, we prompt the model regarding a particular quote without providing any additional detail about the quote or its author. 

\textbf{RAG pipeline:} In the evidence-conditioned setting, we implement a RAG pipeline in which each query quote is paired with semantically similar examples from a retrieval corpus. Specifically, we compute embeddings for all quotes in the dataset using a pretrained embedding model (OpenAI's \texttt{text-embedding-3-small} \citep{openai_embeddings_2024}) and perform nearest-neighbor search to retrieve the top-$k$ most similar quotes with $k=5$ based on cosine similarity, along with their corresponding authors. The query quote-author pair is always retrieved with a similarity score of $1.0$, along with the four other most similar quotes. These five quote–author pairs and their similarity scores are appended to the prompt.

\textbf{Identifying authors:} For each prompt, we identify whether the correct, wrong, or no author was provided by the LLM. We identify author mentions in model outputs using regex-based heuristics. For $M(q)$, we detect whether any author-like name appears in the output without restricting to authors in the dataset. To determine whether $a(q) \in M(q)$, we match against the ground-truth author name and known spelling or alias variants using case-insensitive string matching. We evaluated a subset of 1,000 random classifications using LLM-as-a-judge with \texttt{GPT-4o-mini} \citep{openai2024gpt4ocard} and report 99\% accuracy.

\subsection{Experimental results}

\begin{figure}[t]
    \centering
    \vspace{-10pt} 
    \includegraphics[width=\linewidth]{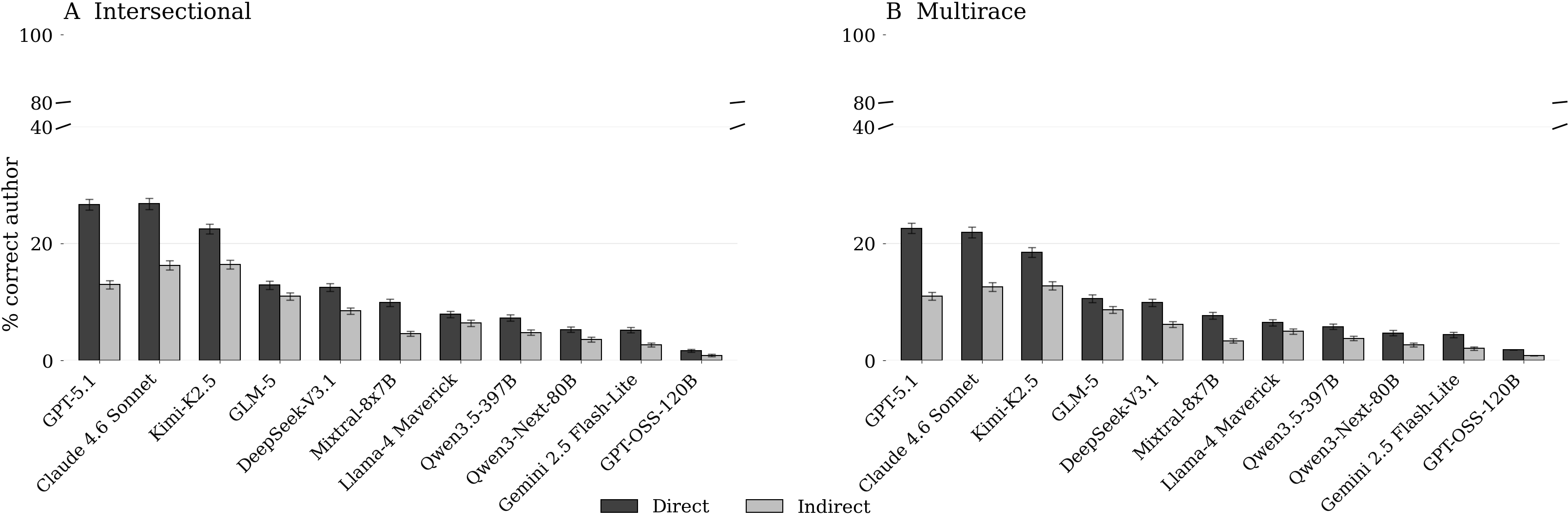}
    \caption{
    Overall attribution accuracy (\% correct) across models and prompts. 
    Note remarkably low performance even on frontier models. Error bars show 95\% confidence intervals.
    }
    \label{fig:overall_accuracy}
\end{figure}

\textbf{Overall attribution accuracy is low.} Overall attribution accuracy is low across all evaluated models and settings, with even the strongest models achieving modest performance. Under direct prompting, where we explicitly ask the model to provide authorship information, frontier models (e.g. \texttt{GPT-5.1}, \texttt{Claude 4.6 Sonnet}) achieve $\sim$25-27\% accuracy on the intersectional dataset, and $\sim$21-23\% accuracy on the multirace dataset. Models such as \texttt{Mixtral-8x7B, Llama-4 Maverick, Qwen3.5-397B, Qwen3-Next-80B,} and \texttt{Gemini 2.5 Flash-Lite} all achieve under 10\% accuracy on both datasets. Interestingly, the model ranking of direct and indirect accuracy performance is distinct: e.g., \texttt{GPT-5.1} outperforms \texttt{Kimi-K2.5} under direct prompting (26.7\% vs. 22.5\% intersectional; 22.6\% vs. 18.5\% multirace), but is surpassed under indirect prompting (13.0\% vs. 16.4\%; 11.0\% vs. 12.8\%). This suggests that a model possessing the correct attribution does not reliably translate into expressing it in more open-ended scenarios, and attribution knowledge is often latent. Results are shown in Figure~\ref{fig:overall_accuracy}; results in the RAG setting are shown in Appendix~\ref{sec:rag-accuracy}.
\textbf{Takeaway: quote attribution remains a difficult task for even state-of-the-art LLMs.}

\begin{figure}[t]
    \centering
    \vspace{-10pt} 
    \includegraphics[width=\linewidth]{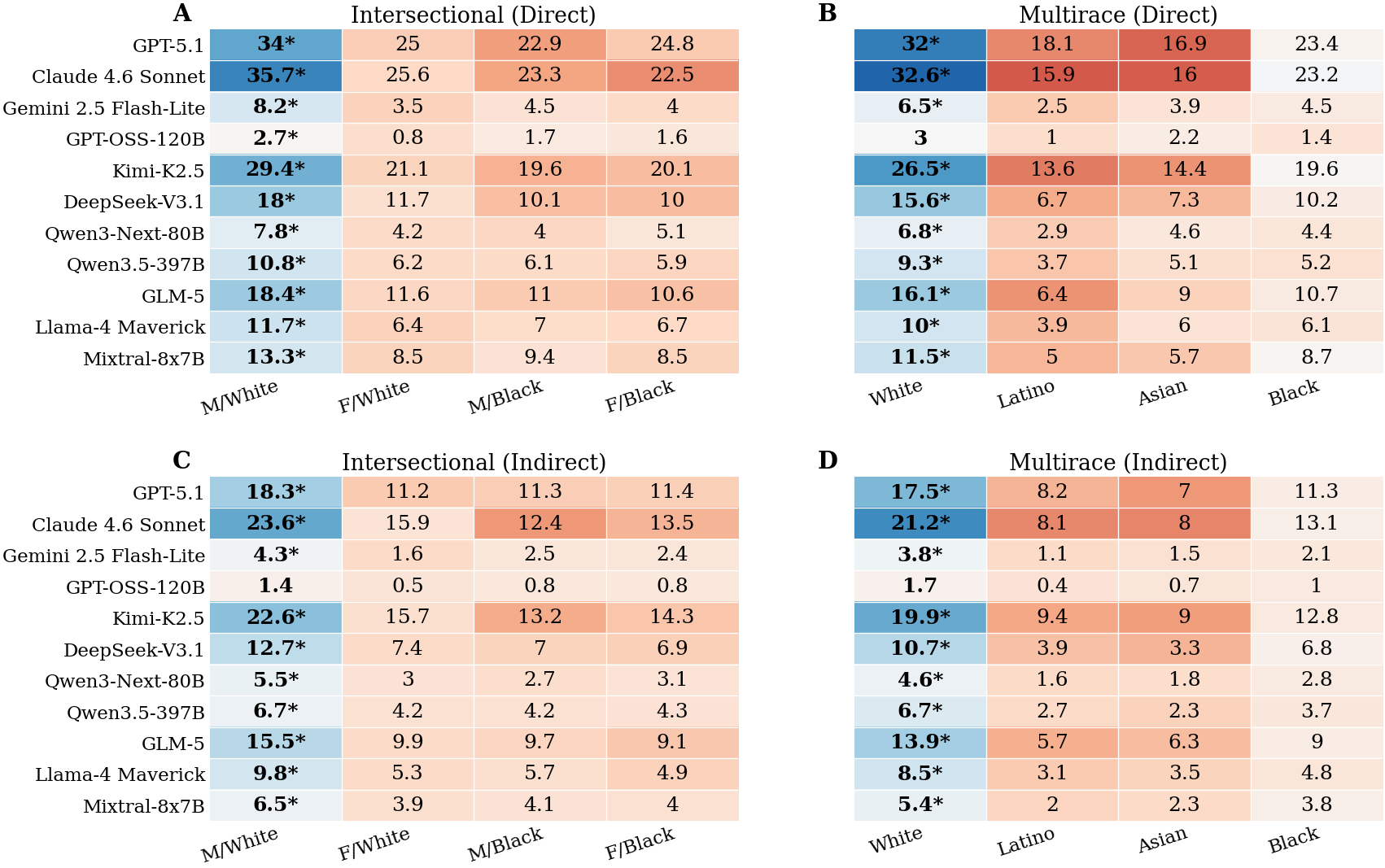}
    \caption{
    Subgroup-level quote attribution accuracy (\% correct author) across models. Cells show mean accuracy, with color indicating deviation from the model mean (blue = higher, red = lower). Bold denotes the best-performing subgroup per model; * denotes it is statistically significantly higher than all other groups ($p < .05$). Across 10 out of 11 models, accuracy is significantly highest for White male and White subgroups.}
    \label{fig:accuracy_across_groups}
\end{figure}

\textbf{There are systematic attribution disparities across demographic subgroups.} Our analyses disaggregated by demographic groups reveal consistent, statistically significant disparities in attribution performance across subgroups, shown in Figure~\ref{fig:accuracy_across_groups}. In the intersectional dataset, the White male subgroup has statistically significantly higher accuracy than every other subgroup with every model and prompt. With \texttt{GPT-5.1} and \texttt{Claude 4.6 Sonnet} models, direct prompt accuracy for the White male subgroup is about 10\% higher than any other subgroup. With the multirace dataset, White is the statistically significant highest-accuracy subgroup in every model and prompt, except for \texttt{GPT-OSS-120B}, but this is likely due to the fact that this is the lowest performing model (1.7\% direct-prompt overall accuracy on intersectional and 1.9\% on multirace). With \texttt{GPT-5.1} and \texttt{Claude 4.6 Sonnet}, direct prompt accuracy for the White subgroup is about 10\% higher than the Black subgroup. For 9 out of 11 models, White subgroup direct prompt accuracy is at least twice the direct-prompt accuracy of both the Latino and Asian subgroups. Black female authors consistently exhibit the lowest accuracy in the intersectional setting with both direct and indirect prompts. 
\noindent\textbf{Takeaway: 
White authors, particularly men, are consistently attributed correctly at significantly higher rates than all other groups, across all models.}

\begin{figure}[t]
    \centering
    \includegraphics[width=\linewidth]{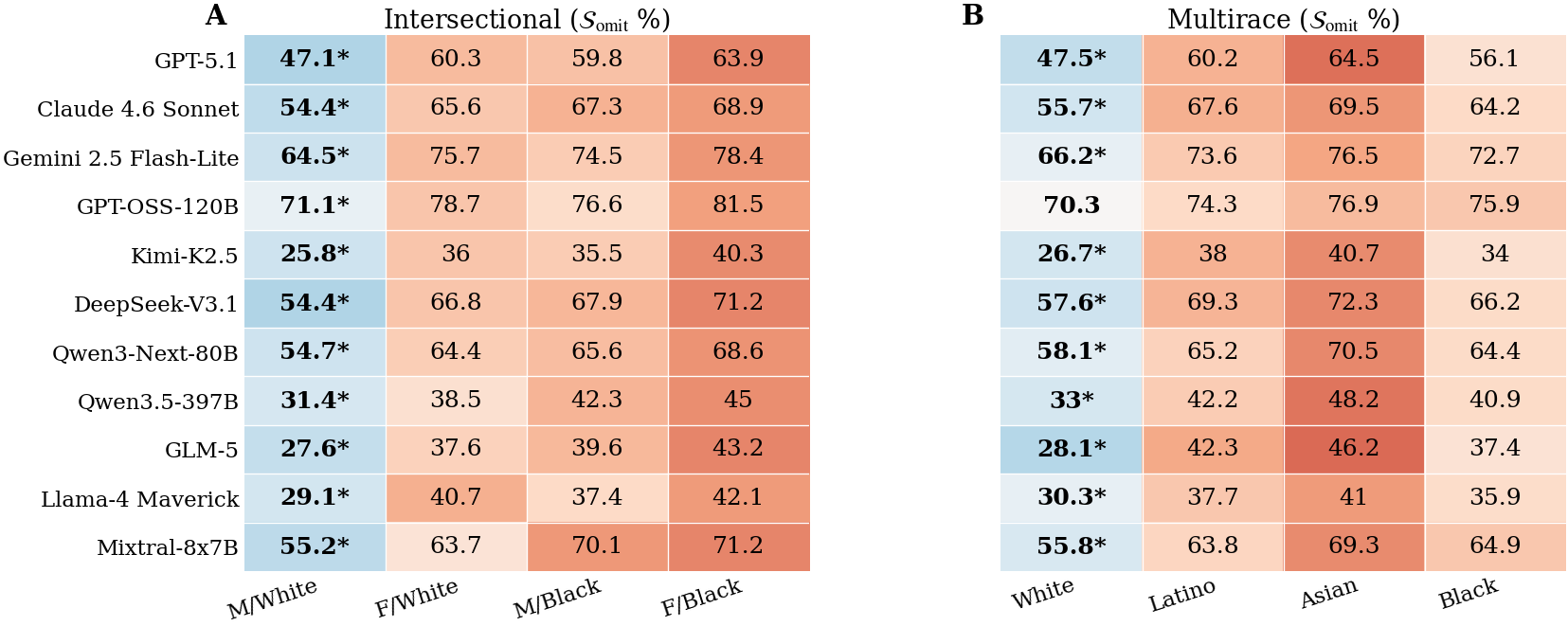}
    \caption{
    Omission suppression $\mathcal{S}_{\mathrm{omit}}$: probability of producing no author under indirect prompting without evidence. Cells show mean suppression (\%), with color indicating deviation from the model mean (blue = lower, red = higher). Bold denotes the lowest suppression subgroup per model; * denotes it is statistically significantly lower than all other groups ($p < .05$). Across models, suppression is consistently lowest for White male (intersectional) and White (multirace) subgroups.
    }
    \label{fig:s_omit}
\end{figure}

\begin{figure}[t]
    \centering
    \vspace{-10pt} 
    \includegraphics[width=\linewidth]{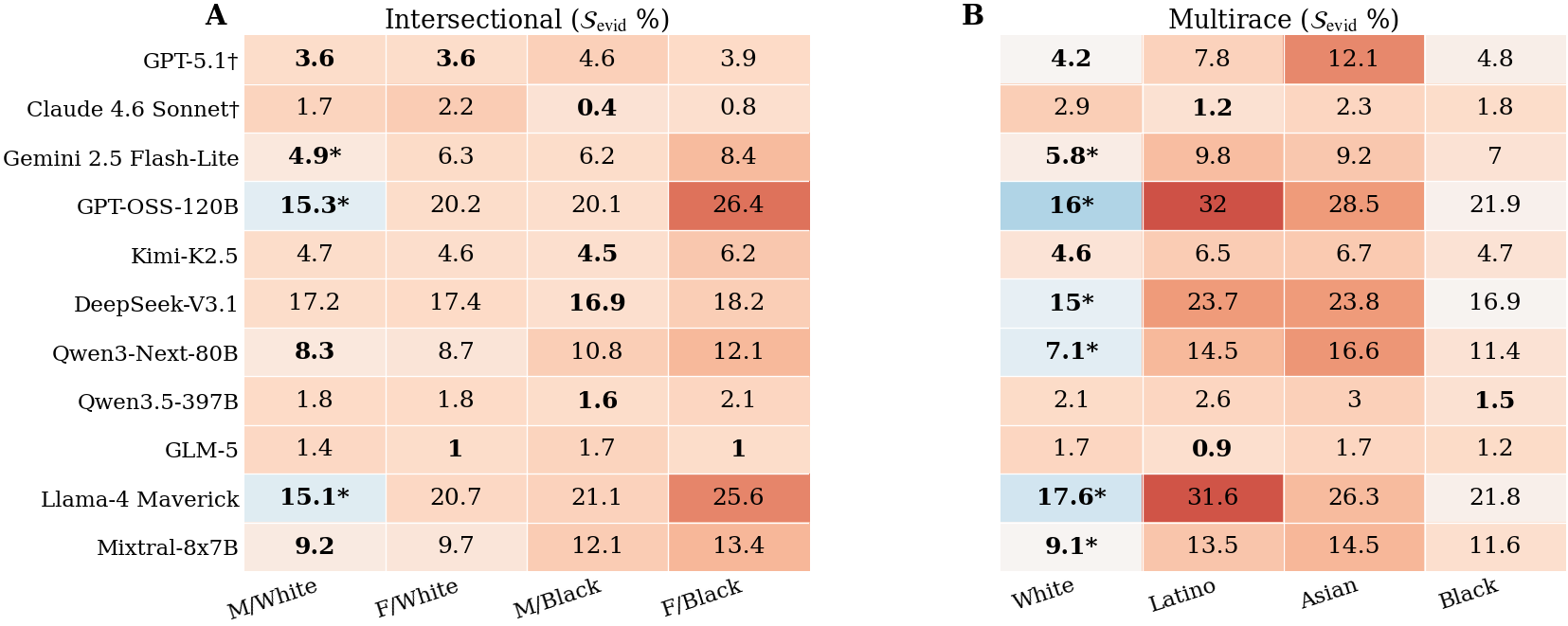}
    \caption{
    Evidence-conditioned suppression $\mathcal{S}_{\mathrm{evid}}$: probability of failing to produce the correct author under indirect prompting when the correct author is explicitly present in the input. Cells show mean suppression (\%), with color indicating deviation from the model mean (blue = lower, red = higher). Bold marks the lowest suppression subgroup per model; * denotes it is statistically significantly lower than all other subgroups ($p < .05$).
    }
    \label{fig:s_evid}
\end{figure}

\textbf{Suppression is a distinct failure mode with systematic disparities across groups.} We now analyze \textit{suppression}, a novel and distinct failure mode not captured by standard accuracy and fairness metrics. Accuracy does not distinguish between incorrect attribution and complete attribution omission, motivating this separate analysis. Suppression captures cases where model discretion determines when to attribute outside of a direct instruction. Through this metric, we do not claim that there is necessarily a \textit{correct} answer between providing author attribution and omitting attribution, but rather measure whether there are disparities in this decision correlated with race and gender. An example of attribution versus suppression is shown in Figure~\ref{fig:example_of_suppression}. We measure suppression with two metrics: $\mathcal{S}_{\mathrm{omit}}$ and $\mathcal{S}_{\mathrm{evid}}$. For $\mathcal{S}_{\mathrm{omit}}$, we measure the absence of any author mention with the indirect prompt and no evidence setting; high $\mathcal{S}_{\mathrm{omit}}$ indicates more omission.  The results are shown in Figure~\ref{fig:s_omit}. For $\mathcal{S}_{\mathrm{evid}}$, we measure the absence of the \textit{correct} author mention with the indirect prompt and evidence-conditioned setting; high $\mathcal{S}_{\mathrm{evid}}$ indicates more incorrect or absent attributions. The results are shown in Figure~\ref{fig:s_evid}. Across models and settings, we observe that suppression is systematically higher for certain demographic groups, indicating that attribution failures manifest as structured patterns of omission. Appendix~\ref{sec:direct-omission} shows author omission rates for the direct prompt, though we note this as task failure, rather than suppression.

We observe that White (and White male) authors exhibit the lowest omission suppression in every model. All other subgroups exhibit statistically significantly higher $\mathcal{S}_{\mathrm{omit}}$ (except for \texttt{GPT-OSS-120B}). $\mathcal{S}_{\mathrm{omit}}$ for White males is on average 10 percentage points lower than for Black males and White females, and 15 points lower than for Black females. $\mathcal{S}_{\mathrm{omit}}$ for White authors is on average 10 percentage points lower than for Latino and Asian authors, and 8 points lower than for Black authors. This trend persists in the evidence-conditioned setting, indicating that models fail to use available information uniformly across groups. Compared to $\mathcal{S}_{\mathrm{omit}}$, subgroup differences are less sharply separated as defined by $\mathcal{S}_{\mathrm{evid}}$, indicating that providing evidence reduces but does not eliminate suppression disparities. In this setting, White male and White subgroups are the minimum suppression subgroups or statistically indistinguishable from it in every model in the intersectional case and in 10 out of 11 models in the multirace case. The White subgroup reports, on average, 6\% less  $\mathcal{S}_{\mathrm{evid}}$ than Latino and Asian subgroups and the White male subgroup reports, on average, 4\% less  $\mathcal{S}_{\mathrm{evid}}$ than the Black female subgroup. We repeat our experiments over related prompts, finding consistent results, as described in Appendix~\ref{sec:indirect-overt-prompting}. \textbf{Takeaway: 
suppression reveals structured disparities across models in whose authorship is acknowledged, again favoring White (male) authors. This
highlights a distinct axis of representational unfairness.}

\section{Conclusion}
We introduce \textsc{AttriBench}, the first fame-controlled, demographically annotated quote attribution dataset to date, and benchmark it on several frontier LLMs. Further, we introduce  \textit{suppression} as a distinct failure mode, capturing selective authorship attribution in LLMs. We find that attribution remains a challenging task for LLMs and displays systematic disparities: models are more accurate and less likely to suppress White, particularly White male, authors.  
Our work opens the door for further exploration of bias in open-ended attribution: our evaluation does not capture attribution in online settings, nor does it perfectly capture fame or other confounding variables impacting attribution. We look forward to this future work, and believe that our findings position quote attribution as a novel setting for studying representational fairness, highlighting how LLMs mediate whose authorship is recognized.

\section*{Acknowledgments}

We thank NYU HPC for their generous support and help in computational simulations and experiments. This material is based upon work supported by the National Science Foundation Graduate Research Fellowship Program under Grant No. DGE-2234660. Any opinions, findings, and conclusions or recommendations expressed in this material are those of the author(s) and do not necessarily reflect the views of the National Science Foundation. This work was supported by the Institute of Information \& Communications Technology Planning \& Evaluation (IITP) with a grant funded by the Ministry of Science and ICT (MSIT) of the Republic of Korea in connection with the Global AI Frontier Lab International Collaborative Research (No. RS-2024-00509258).

\section*{Ethics Statement}

The goal of our work is to promote ethical evaluation of LLMs by identifying and quantifying representational disparities in attribution. We aim for our benchmark dataset to support more equitable attribution in LLMs.

\section*{Reproducibility Statement}

The code and data used for our experiments and analyses are publicly available at \url{https://github.com/elizaberman/attribution-bias-in-llms}. We evaluated a combination of proprietary and open-weight models accessed through provider APIs, Hugging Face, and Together AI. Exact model names, versions, and access methods are provided in Appendix~\ref{sec:appendix_models}.

\bibliographystyle{colm2026_conference}
\bibliography{colm2026_conference}

@inproceedings{Dhamala_2021, series={FAccT ’21},
   title={BOLD: Dataset and Metrics for Measuring Biases in Open-Ended Language Generation},
   url={http://dx.doi.org/10.1145/3442188.3445924},
   DOI={10.1145/3442188.3445924},
   booktitle={Proceedings of the 2021 ACM Conference on Fairness, Accountability, and Transparency},
   publisher={ACM},
   author={Dhamala, Jwala and Sun, Tony and Kumar, Varun and Krishna, Satyapriya and Pruksachatkun, Yada and Chang, Kai-Wei and Gupta, Rahul},
   year={2021},
   month=mar, pages={862–872},
   collection={FAccT ’21} }

@misc{nadeem2020stereosetmeasuringstereotypicalbias,
      title={StereoSet: Measuring stereotypical bias in pretrained language models}, 
      author={Moin Nadeem and Anna Bethke and Siva Reddy},
      year={2020},
      eprint={2004.09456},
      archivePrefix={arXiv},
      primaryClass={cs.CL},
      url={https://arxiv.org/abs/2004.09456}, 
}

@misc{parrish2022bbqhandbuiltbiasbenchmark,
      title={BBQ: A Hand-Built Bias Benchmark for Question Answering}, 
      author={Alicia Parrish and Angelica Chen and Nikita Nangia and Vishakh Padmakumar and Jason Phang and Jana Thompson and Phu Mon Htut and Samuel R. Bowman},
      year={2022},
      eprint={2110.08193},
      archivePrefix={arXiv},
      primaryClass={cs.CL},
      url={https://arxiv.org/abs/2110.08193}, 
}

@misc{press2024citemelanguagemodelsaccurately,
      title={CiteME: Can Language Models Accurately Cite Scientific Claims?}, 
      author={Ori Press and Andreas Hochlehnert and Ameya Prabhu and Vishaal Udandarao and Ofir Press and Matthias Bethge},
      year={2024},
      eprint={2407.12861},
      archivePrefix={arXiv},
      primaryClass={cs.CL},
      url={https://arxiv.org/abs/2407.12861}, 
}

@inproceedings{zhong-etal-2024-said,
    title = "Who Said What: Formalization and Benchmarks for the Task of Quote Attribution",
    author = "Zhong, Wenjie  and
      Naradowsky, Jason  and
      Takamura, Hiroya  and
      Kobayashi, Ichiro  and
      Miyao, Yusuke",
    editor = "Calzolari, Nicoletta  and
      Kan, Min-Yen  and
      Hoste, Veronique  and
      Lenci, Alessandro  and
      Sakti, Sakriani  and
      Xue, Nianwen",
    booktitle = "Proceedings of the 2024 Joint International Conference on Computational Linguistics, Language Resources and Evaluation (LREC-COLING 2024)",
    month = may,
    year = "2024",
    address = "Torino, Italia",
    publisher = "ELRA and ICCL",
    url = "https://aclanthology.org/2024.lrec-main.1530/",
    pages = "17588--17602",
    }

@misc{he2025getscitedgendermajoritybias,
      title={Who Gets Cited? Gender- and Majority-Bias in LLM-Driven Reference Selection}, 
      author={Jiangen He},
      year={2025},
      eprint={2508.02740},
      archivePrefix={arXiv},
      primaryClass={cs.DL},
      url={https://arxiv.org/abs/2508.02740}, 
}

@inproceedings{goel2018proposing,
author = {Shivali Goel and Rishi Madhok and Shweta Garg},
title = {Proposing Contextually Relevant Quotes for Images},
booktitle = {Advances in Information Retrieval},
pages = {591--597},
year = {2018},
publisher = {Springer},
doi = {10.1007/978-3-319-76941-7_49}
}

@misc{dataforseo_api,
author = {{DataForSEO}},
title = {DataForSEO API Documentation},
howpublished = {\url{https://docs.dataforseo.com/}},
}

@misc{kimiteam2026kimik25visualagentic,
      title={Kimi K2.5: Visual Agentic Intelligence}, 
      author={{Kimi Team} and Tongtong Bai and Yifan Bai and Yiping Bao and S. H. Cai and Yuan Cao and Y. Charles and H. S. Che and Cheng Chen and Guanduo Chen and Huarong Chen and Jia Chen and Jiahao Chen and Jianlong Chen and Jun Chen and Kefan Chen and Liang Chen and Ruijue Chen and Xinhao Chen and Yanru Chen and Yanxu Chen and Yicun Chen and Yimin Chen and Yingjiang Chen and Yuankun Chen and Yujie Chen and Yutian Chen and Zhirong Chen and Ziwei Chen and Dazhi Cheng and Minghan Chu and Jialei Cui and Jiaqi Deng and Muxi Diao and Hao Ding and Mengfan Dong and Mengnan Dong and Yuxin Dong and Yuhao Dong and Angang Du and Chenzhuang Du and Dikang Du and Lingxiao Du and Yulun Du and Yu Fan and Shengjun Fang and Qiulin Feng and Yichen Feng and Garimugai Fu and Kelin Fu and Hongcheng Gao and Tong Gao and Yuyao Ge and Shangyi Geng and Chengyang Gong and Xiaochen Gong and Zhuoma Gongque and Qizheng Gu and Xinran Gu and Yicheng Gu and Longyu Guan and Yuanying Guo and Xiaoru Hao and Weiran He and Wenyang He and Yunjia He and Chao Hong and Hao Hu and Jiaxi Hu and Yangyang Hu and Zhenxing Hu and Ke Huang and Ruiyuan Huang and Weixiao Huang and Zhiqi Huang and Tao Jiang and Zhejun Jiang and Xinyi Jin and Yu Jing and Guokun Lai and Aidi Li and C. Li and Cheng Li and Fang Li and Guanghe Li and Guanyu Li and Haitao Li and Haoyang Li and Jia Li and Jingwei Li and Junxiong Li and Lincan Li and Mo Li and Weihong Li and Wentao Li and Xinhang Li and Xinhao Li and Yang Li and Yanhao Li and Yiwei Li and Yuxiao Li and Zhaowei Li and Zheming Li and Weilong Liao and Jiawei Lin and Xiaohan Lin and Zhishan Lin and Zichao Lin and Cheng Liu and Chenyu Liu and Hongzhang Liu and Liang Liu and Shaowei Liu and Shudong Liu and Shuran Liu and Tianwei Liu and Tianyu Liu and Weizhou Liu and Xiangyan Liu and Yangyang Liu and Yanming Liu and Yibo Liu and Yuanxin Liu and Yue Liu and Zhengying Liu and Zhongnuo Liu and Enzhe Lu and Haoyu Lu and Zhiyuan Lu and Junyu Luo and Tongxu Luo and Yashuo Luo and Long Ma and Yingwei Ma and Shaoguang Mao and Yuan Mei and Xin Men and Fanqing Meng and Zhiyong Meng and Yibo Miao and Minqing Ni and Kun Ouyang and Siyuan Pan and Bo Pang and Yuchao Qian and Ruoyu Qin and Zeyu Qin and Jiezhong Qiu and Bowen Qu and Zeyu Shang and Youbo Shao and Tianxiao Shen and Zhennan Shen and Juanfeng Shi and Lidong Shi and Shengyuan Shi and Feifan Song and Pengwei Song and Tianhui Song and Xiaoxi Song and Hongjin Su and Jianlin Su and Zhaochen Su and Lin Sui and Jinsong Sun and Junyao Sun and Tongyu Sun and Flood Sung and Yunpeng Tai and Chuning Tang and Heyi Tang and Xiaojuan Tang and Zhengyang Tang and Jiawen Tao and Shiyuan Teng and Chaoran Tian and Pengfei Tian and Ao Wang and Bowen Wang and Chensi Wang and Chuang Wang and Congcong Wang and Dingkun Wang and Dinglu Wang and Dongliang Wang and Feng Wang and Hailong Wang and Haiming Wang and Hengzhi Wang and Huaqing Wang and Hui Wang and Jiahao Wang and Jinhong Wang and Jiuzheng Wang and Kaixin Wang and Linian Wang and Qibin Wang and Shengjie Wang and Shuyi Wang and Si Wang and Wei Wang and Xiaochen Wang and Xinyuan Wang and Yao Wang and Yejie Wang and Yipu Wang and Yiqin Wang and Yucheng Wang and Yuzhi Wang and Zhaoji Wang and Zhaowei Wang and Zhengtao Wang and Zhexu Wang and Zihan Wang and Zizhe Wang and Chu Wei and Ming Wei and Chuan Wen and Zichen Wen and Chengjie Wu and Haoning Wu and Junyan Wu and Rucong Wu and Wenhao Wu and Yuefeng Wu and Yuhao Wu and Yuxin Wu and Zijian Wu and Chenjun Xiao and Jin Xie and Xiaotong Xie and Yuchong Xie and Yifei Xin and Bowei Xing and Boyu Xu and Jianfan Xu and Jing Xu and Jinjing Xu and L. H. Xu and Lin Xu and Suting Xu and Weixin Xu and Xinbo Xu and Xinran Xu and Yangchuan Xu and Yichang Xu and Yuemeng Xu and Zelai Xu and Ziyao Xu and Junjie Yan and Yuzi Yan and Guangyao Yang and Hao Yang and Junwei Yang and Kai Yang and Ningyuan Yang and Ruihan Yang and Xiaofei Yang and Xinlong Yang and Ying Yang and Yi Yang and Yi Yang and Zhen Yang and Zhilin Yang and Zonghan Yang and Haotian Yao and Dan Ye and Wenjie Ye and Zhuorui Ye and Bohong Yin and Chengzhen Yu and Longhui Yu and Tao Yu and Tianxiang Yu and Enming Yuan and Mengjie Yuan and Xiaokun Yuan and Yang Yue and Weihao Zeng and Dunyuan Zha and Haobing Zhan and Dehao Zhang and Hao Zhang and Jin Zhang and Puqi Zhang and Qiao Zhang and Rui Zhang and Xiaobin Zhang and Y. Zhang and Yadong Zhang and Yangkun Zhang and Yichi Zhang and Yizhi Zhang and Yongting Zhang and Yu Zhang and Yushun Zhang and Yutao Zhang and Yutong Zhang and Zheng Zhang and Chenguang Zhao and Feifan Zhao and Jinxiang Zhao and Shuai Zhao and Xiangyu Zhao and Yikai Zhao and Zijia Zhao and Huabin Zheng and Ruihan Zheng and Shaojie Zheng and Tengyang Zheng and Junfeng Zhong and Longguang Zhong and Weiming Zhong and M. Zhou and Runjie Zhou and Xinyu Zhou and Zaida Zhou and Jinguo Zhu and Liya Zhu and Xinhao Zhu and Yuxuan Zhu and Zhen Zhu and Jingze Zhuang and Weiyu Zhuang and Ying Zou and Xinxing Zu},
      year={2026},
      eprint={2602.02276},
      archivePrefix={arXiv},
      primaryClass={cs.CL},
      url={https://arxiv.org/abs/2602.02276}, 
}

@misc{openai2024gpt4ocard,
      title={GPT-4o System Card}, 
      author={OpenAI and Aaron Hurst and Adam Lerer and Adam P. Goucher and Adam Perelman and Aditya Ramesh and Aidan Clark and AJ Ostrow and Akila Welihinda and Alan Hayes and Alec Radford and Aleksander Mądry and Alex Baker-Whitcomb and Alex Beutel and Alex Borzunov and Alex Carney and Alex Chow and Alex Kirillov and Alex Nichol and Alex Paino and Alex Renzin and Alex Tachard Passos and Alexander Kirillov and Alexi Christakis and Alexis Conneau and Ali Kamali and Allan Jabri and Allison Moyer and Allison Tam and Amadou Crookes and Amin Tootoochian and Amin Tootoonchian and Ananya Kumar and Andrea Vallone and Andrej Karpathy and Andrew Braunstein and Andrew Cann and Andrew Codispoti and Andrew Galu and Andrew Kondrich and Andrew Tulloch and Andrey Mishchenko and Angela Baek and Angela Jiang and Antoine Pelisse and Antonia Woodford and Anuj Gosalia and Arka Dhar and Ashley Pantuliano and Avi Nayak and Avital Oliver and Barret Zoph and Behrooz Ghorbani and Ben Leimberger and Ben Rossen and Ben Sokolowsky and Ben Wang and Benjamin Zweig and Beth Hoover and Blake Samic and Bob McGrew and Bobby Spero and Bogo Giertler and Bowen Cheng and Brad Lightcap and Brandon Walkin and Brendan Quinn and Brian Guarraci and Brian Hsu and Bright Kellogg and Brydon Eastman and Camillo Lugaresi and Carroll Wainwright and Cary Bassin and Cary Hudson and Casey Chu and Chad Nelson and Chak Li and Chan Jun Shern and Channing Conger and Charlotte Barette and Chelsea Voss and Chen Ding and Cheng Lu and Chong Zhang and Chris Beaumont and Chris Hallacy and Chris Koch and Christian Gibson and Christina Kim and Christine Choi and Christine McLeavey and Christopher Hesse and Claudia Fischer and Clemens Winter and Coley Czarnecki and Colin Jarvis and Colin Wei and Constantin Koumouzelis and Dane Sherburn and Daniel Kappler and Daniel Levin and Daniel Levy and David Carr and David Farhi and David Mely and David Robinson and David Sasaki and Denny Jin and Dev Valladares and Dimitris Tsipras and Doug Li and Duc Phong Nguyen and Duncan Findlay and Edede Oiwoh and Edmund Wong and Ehsan Asdar and Elizabeth Proehl and Elizabeth Yang and Eric Antonow and Eric Kramer and Eric Peterson and Eric Sigler and Eric Wallace and Eugene Brevdo and Evan Mays and Farzad Khorasani and Felipe Petroski Such and Filippo Raso and Francis Zhang and Fred von Lohmann and Freddie Sulit and Gabriel Goh and Gene Oden and Geoff Salmon and Giulio Starace and Greg Brockman and Hadi Salman and Haiming Bao and Haitang Hu and Hannah Wong and Haoyu Wang and Heather Schmidt and Heather Whitney and Heewoo Jun and Hendrik Kirchner and Henrique Ponde de Oliveira Pinto and Hongyu Ren and Huiwen Chang and Hyung Won Chung and Ian Kivlichan and Ian O'Connell and Ian O'Connell and Ian Osband and Ian Silber and Ian Sohl and Ibrahim Okuyucu and Ikai Lan and Ilya Kostrikov and Ilya Sutskever and Ingmar Kanitscheider and Ishaan Gulrajani and Jacob Coxon and Jacob Menick and Jakub Pachocki and James Aung and James Betker and James Crooks and James Lennon and Jamie Kiros and Jan Leike and Jane Park and Jason Kwon and Jason Phang and Jason Teplitz and Jason Wei and Jason Wolfe and Jay Chen and Jeff Harris and Jenia Varavva and Jessica Gan Lee and Jessica Shieh and Ji Lin and Jiahui Yu and Jiayi Weng and Jie Tang and Jieqi Yu and Joanne Jang and Joaquin Quinonero Candela and Joe Beutler and Joe Landers and Joel Parish and Johannes Heidecke and John Schulman and Jonathan Lachman and Jonathan McKay and Jonathan Uesato and Jonathan Ward and Jong Wook Kim and Joost Huizinga and Jordan Sitkin and Jos Kraaijeveld and Josh Gross and Josh Kaplan and Josh Snyder and Joshua Achiam and Joy Jiao and Joyce Lee and Juntang Zhuang and Justyn Harriman and Kai Fricke and Kai Hayashi and Karan Singhal and Katy Shi and Kavin Karthik and Kayla Wood and Kendra Rimbach and Kenny Hsu and Kenny Nguyen and Keren Gu-Lemberg and Kevin Button and Kevin Liu and Kiel Howe and Krithika Muthukumar and Kyle Luther and Lama Ahmad and Larry Kai and Lauren Itow and Lauren Workman and Leher Pathak and Leo Chen and Li Jing and Lia Guy and Liam Fedus and Liang Zhou and Lien Mamitsuka and Lilian Weng and Lindsay McCallum and Lindsey Held and Long Ouyang and Louis Feuvrier and Lu Zhang and Lukas Kondraciuk and Lukasz Kaiser and Luke Hewitt and Luke Metz and Lyric Doshi and Mada Aflak and Maddie Simens and Madelaine Boyd and Madeleine Thompson and Marat Dukhan and Mark Chen and Mark Gray and Mark Hudnall and Marvin Zhang and Marwan Aljubeh and Mateusz Litwin and Matthew Zeng and Max Johnson and Maya Shetty and Mayank Gupta and Meghan Shah and Mehmet Yatbaz and Meng Jia Yang and Mengchao Zhong and Mia Glaese and Mianna Chen and Michael Janner and Michael Lampe and Michael Petrov and Michael Wu and Michele Wang and Michelle Fradin and Michelle Pokrass and Miguel Castro and Miguel Oom Temudo de Castro and Mikhail Pavlov and Miles Brundage and Miles Wang and Minal Khan and Mira Murati and Mo Bavarian and Molly Lin and Murat Yesildal and Nacho Soto and Natalia Gimelshein and Natalie Cone and Natalie Staudacher and Natalie Summers and Natan LaFontaine and Neil Chowdhury and Nick Ryder and Nick Stathas and Nick Turley and Nik Tezak and Niko Felix and Nithanth Kudige and Nitish Keskar and Noah Deutsch and Noel Bundick and Nora Puckett and Ofir Nachum and Ola Okelola and Oleg Boiko and Oleg Murk and Oliver Jaffe and Olivia Watkins and Olivier Godement and Owen Campbell-Moore and Patrick Chao and Paul McMillan and Pavel Belov and Peng Su and Peter Bak and Peter Bakkum and Peter Deng and Peter Dolan and Peter Hoeschele and Peter Welinder and Phil Tillet and Philip Pronin and Philippe Tillet and Prafulla Dhariwal and Qiming Yuan and Rachel Dias and Rachel Lim and Rahul Arora and Rajan Troll and Randall Lin and Rapha Gontijo Lopes and Raul Puri and Reah Miyara and Reimar Leike and Renaud Gaubert and Reza Zamani and Ricky Wang and Rob Donnelly and Rob Honsby and Rocky Smith and Rohan Sahai and Rohit Ramchandani and Romain Huet and Rory Carmichael and Rowan Zellers and Roy Chen and Ruby Chen and Ruslan Nigmatullin and Ryan Cheu and Saachi Jain and Sam Altman and Sam Schoenholz and Sam Toizer and Samuel Miserendino and Sandhini Agarwal and Sara Culver and Scott Ethersmith and Scott Gray and Sean Grove and Sean Metzger and Shamez Hermani and Shantanu Jain and Shengjia Zhao and Sherwin Wu and Shino Jomoto and Shirong Wu and Shuaiqi and Xia and Sonia Phene and Spencer Papay and Srinivas Narayanan and Steve Coffey and Steve Lee and Stewart Hall and Suchir Balaji and Tal Broda and Tal Stramer and Tao Xu and Tarun Gogineni and Taya Christianson and Ted Sanders and Tejal Patwardhan and Thomas Cunninghman and Thomas Degry and Thomas Dimson and Thomas Raoux and Thomas Shadwell and Tianhao Zheng and Todd Underwood and Todor Markov and Toki Sherbakov and Tom Rubin and Tom Stasi and Tomer Kaftan and Tristan Heywood and Troy Peterson and Tyce Walters and Tyna Eloundou and Valerie Qi and Veit Moeller and Vinnie Monaco and Vishal Kuo and Vlad Fomenko and Wayne Chang and Weiyi Zheng and Wenda Zhou and Wesam Manassra and Will Sheu and Wojciech Zaremba and Yash Patil and Yilei Qian and Yongjik Kim and Youlong Cheng and Yu Zhang and Yuchen He and Yuchen Zhang and Yujia Jin and Yunxing Dai and Yury Malkov},
      year={2024},
      eprint={2410.21276},
      archivePrefix={arXiv},
      primaryClass={cs.CL},
      url={https://arxiv.org/abs/2410.21276}, 
}

@misc{wolf2020huggingfacestransformersstateoftheartnatural,
      title={HuggingFace's Transformers: State-of-the-art Natural Language Processing}, 
      author={Thomas Wolf and Lysandre Debut and Victor Sanh and Julien Chaumond and Clement Delangue and Anthony Moi and Pierric Cistac and Tim Rault and Rémi Louf and Morgan Funtowicz and Joe Davison and Sam Shleifer and Patrick von Platen and Clara Ma and Yacine Jernite and Julien Plu and Canwen Xu and Teven Le Scao and Sylvain Gugger and Mariama Drame and Quentin Lhoest and Alexander M. Rush},
      year={2020},
      eprint={1910.03771},
      archivePrefix={arXiv},
      primaryClass={cs.CL},
      url={https://arxiv.org/abs/1910.03771}, 
}

@misc{openai_embeddings_2024,
  title = {New embedding models and API updates},
  author = {{OpenAI}},
  year = {2024},
  howpublished = {\url{https://openai.com/blog/new-embedding-models-and-api-updates}}
}

@misc{anthropic2026claude46sonnet,
  author       = {{Anthropic}},
  title        = {Claude Sonnet 4.6 System Card},
  year         = {2026},
  url          = {https://www.anthropic.com/claude/sonnet},
}

@misc{google2025gemini25flashlite,
  author       = {{Google DeepMind}},
  title        = {Gemini 2.5 Flash-Lite Model Documentation},
  year         = {2025},
  url          = {https://ai.google.dev/gemini-api/docs/models/gemini-2.5-flash-lite},
}

@misc{openai2025gpt51,
  author       = {{OpenAI}},
  title        = {GPT-5.1 Instant and GPT-5.1 Thinking System Card Addendum},
  year         = {2025},
  month        = nov,
  howpublished = {\url{https://openai.com/index/gpt-5-system-card-addendum-gpt-5-1/}},
}

@techreport{perplexity_sonar_default_2024,
  title       = {Sonar: Search-Augmented Language Models},
  author      = {{Perplexity AI}},
  institution = {Perplexity AI},
  year        = {2024},
  url         = {https://docs.perplexity.ai/docs/sonar-models},
  note        = {Default Sonar model}
}

@misc{wikidata_rest_api,
  author       = {{Wikidata}},
  title        = {Wikidata: REST API},
  year         = {2026},
  url          = {https://www.wikidata.org/wiki/Wikidata:REST_API},
}

@inproceedings{michel-etal-2025-evaluating,
    title = "Evaluating {LLM}s for Quotation Attribution in Literary Texts: A Case Study of {LL}a{M}a3",
    author = "Michel, Gaspard  and
      Epure, Elena V.  and
      Hennequin, Romain  and
      Cerisara, Christophe",
    editor = "Chiruzzo, Luis  and
      Ritter, Alan  and
      Wang, Lu",
    booktitle = "Proceedings of the 2025 Conference of the Nations of the Americas Chapter of the Association for Computational Linguistics: Human Language Technologies (Volume 2: Short Papers)",
    month = apr,
    year = "2025",
    address = "Albuquerque, New Mexico",
    publisher = "Association for Computational Linguistics",
    url = "https://aclanthology.org/2025.naacl-short.62/",
    doi = "10.18653/v1/2025.naacl-short.62",
    pages = "742--755",
    ISBN = "979-8-89176-190-2",
}

@inproceedings{vishnubhotla-etal-2023-improving,
    title = "Improving Automatic Quotation Attribution in Literary Novels",
    author = "Vishnubhotla, Krishnapriya  and
      Rudzicz, Frank  and
      Hirst, Graeme  and
      Hammond, Adam",
    editor = "Rogers, Anna  and
      Boyd-Graber, Jordan  and
      Okazaki, Naoaki",
    booktitle = "Proceedings of the 61st Annual Meeting of the Association for Computational Linguistics (Volume 2: Short Papers)",
    month = jul,
    year = "2023",
    address = "Toronto, Canada",
    publisher = "Association for Computational Linguistics",
    url = "https://aclanthology.org/2023.acl-short.64/",
    doi = "10.18653/v1/2023.acl-short.64",
    pages = "737--746",
}

@inproceedings{vishnubhotla-etal-2022-project,
    title = "The Project Dialogism Novel Corpus: A Dataset for Quotation Attribution in Literary Texts",
    author = "Vishnubhotla, Krishnapriya  and
      Hammond, Adam  and
      Hirst, Graeme",
    editor = "Calzolari, Nicoletta  and
      B{\'e}chet, Fr{\'e}d{\'e}ric  and
      Blache, Philippe  and
      Choukri, Khalid  and
      Cieri, Christopher  and
      Declerck, Thierry  and
      Goggi, Sara  and
      Isahara, Hitoshi  and
      Maegaard, Bente  and
      Mariani, Joseph  and
      Mazo, H{\'e}l{\`e}ne  and
      Odijk, Jan  and
      Piperidis, Stelios",
    booktitle = "Proceedings of the Thirteenth Language Resources and Evaluation Conference",
    month = jun,
    year = "2022",
    address = "Marseille, France",
    publisher = "European Language Resources Association",
    url = "https://aclanthology.org/2022.lrec-1.628/",
    pages = "5838--5848",
}

@inproceedings{10.1145/3437963.3441760,
author = {Vaucher, Timot\'{e} and Spitz, Andreas and Catasta, Michele and West, Robert},
title = {Quotebank: A Corpus of Quotations from a Decade of News},
year = {2021},
isbn = {9781450382977},
publisher = {Association for Computing Machinery},
address = {New York, NY, USA},
url = {https://doi.org/10.1145/3437963.3441760},
doi = {10.1145/3437963.3441760},
booktitle = {Proceedings of the 14th ACM International Conference on Web Search and Data Mining},
pages = {328–336},
numpages = {9},
location = {Virtual Event, Israel},
series = {WSDM '21}
}

@inproceedings{zhang-liu-2022-directquote,
    title = "{D}irect{Q}uote: A Dataset for Direct Quotation Extraction and Attribution in News Articles",
    author = "Zhang, Yuanchi  and
      Liu, Yang",
    editor = "Calzolari, Nicoletta  and
      B{\'e}chet, Fr{\'e}d{\'e}ric  and
      Blache, Philippe  and
      Choukri, Khalid  and
      Cieri, Christopher  and
      Declerck, Thierry  and
      Goggi, Sara  and
      Isahara, Hitoshi  and
      Maegaard, Bente  and
      Mariani, Joseph  and
      Mazo, H{\'e}l{\`e}ne  and
      Odijk, Jan  and
      Piperidis, Stelios",
    booktitle = "Proceedings of the Thirteenth Language Resources and Evaluation Conference",
    month = jun,
    year = "2022",
    address = "Marseille, France",
    publisher = "European Language Resources Association",
    url = "https://aclanthology.org/2022.lrec-1.752/",
    pages = "6959--6966",
}

@inproceedings{abolghasemi-etal-2025-evaluation,
    title = "Evaluation of Attribution Bias in Generator-Aware Retrieval-Augmented Large Language Models",
    author = "Abolghasemi, Amin  and
      Azzopardi, Leif  and
      Hashemi, Seyyed Hadi  and
      de Rijke, Maarten  and
      Verberne, Suzan",
    editor = "Che, Wanxiang  and
      Nabende, Joyce  and
      Shutova, Ekaterina  and
      Pilehvar, Mohammad Taher",
    booktitle = "Findings of the Association for Computational Linguistics: ACL 2025",
    month = jul,
    year = "2025",
    address = "Vienna, Austria",
    publisher = "Association for Computational Linguistics",
    url = "https://aclanthology.org/2025.findings-acl.1087/",
    doi = "10.18653/v1/2025.findings-acl.1087",
    pages = "21105--21124",
    ISBN = "979-8-89176-256-5",
}

@misc{alipoormolabashi2025quantifyingmisattributionunfairnessauthorship,
      title={Quantifying Misattribution Unfairness in Authorship Attribution}, 
      author={Pegah Alipoormolabashi and Ajay Patel and Niranjan Balasubramanian},
      year={2025},
      eprint={2506.02321},
      archivePrefix={arXiv},
      primaryClass={cs.CL},
      url={https://arxiv.org/abs/2506.02321}, 
}

@misc{rashkin2022measuringattributionnaturallanguage,
      title={Measuring Attribution in Natural Language Generation Models}, 
      author={Hannah Rashkin and Vitaly Nikolaev and Matthew Lamm and Lora Aroyo and Michael Collins and Dipanjan Das and Slav Petrov and Gaurav Singh Tomar and Iulia Turc and David Reitter},
      year={2022},
      eprint={2112.12870},
      archivePrefix={arXiv},
      primaryClass={cs.CL},
      url={https://arxiv.org/abs/2112.12870}, 
}

@article{10.1093/jamia/ocaf063,
    author = {Scherbakov, Dmitry and Hubig, Nina and Jansari, Vinita and Bakumenko, Alexander and Lenert, Leslie A},
    title = {The emergence of large language models as tools in literature reviews: a large language model-assisted systematic review},
    journal = {Journal of the American Medical Informatics Association},
    volume = {32},
    number = {6},
    pages = {1071-1086},
    year = {2025},
    month = {05},
    issn = {1527-974X},
    doi = {10.1093/jamia/ocaf063},
    url = {https://doi.org/10.1093/jamia/ocaf063},
    eprint = {https://academic.oup.com/jamia/article-pdf/32/6/1071/63100940/ocaf063.pdf},
}

@inproceedings{10.1145/3613904.3641917,
author = {Wang, Jiyao and Hu, Haolong and Wang, Zuyuan and Yan, Song and Sheng, Youyu and He, Dengbo},
title = {Evaluating Large Language Models on Academic Literature Understanding and Review: An Empirical Study among Early-stage Scholars},
year = {2024},
isbn = {9798400703300},
publisher = {Association for Computing Machinery},
address = {New York, NY, USA},
url = {https://doi.org/10.1145/3613904.3641917},
doi = {10.1145/3613904.3641917},
booktitle = {Proceedings of the 2024 CHI Conference on Human Factors in Computing Systems},
articleno = {12},
numpages = {18},
location = {Honolulu, HI, USA},
series = {CHI '24}
}

@misc{wang2024largelanguagemodelseducation,
      title={Large Language Models for Education: A Survey and Outlook}, 
      author={Shen Wang and Tianlong Xu and Hang Li and Chaoli Zhang and Joleen Liang and Jiliang Tang and Philip S. Yu and Qingsong Wen},
      year={2024},
      eprint={2403.18105},
      archivePrefix={arXiv},
      primaryClass={cs.CL},
      url={https://arxiv.org/abs/2403.18105}, 
}

@misc{liao2024llmsresearchtoolslarge,
      title={LLMs as Research Tools: A Large Scale Survey of Researchers' Usage and Perceptions}, 
      author={Zhehui Liao and Maria Antoniak and Inyoung Cheong and Evie Yu-Yen Cheng and Ai-Heng Lee and Kyle Lo and Joseph Chee Chang and Amy X. Zhang},
      year={2024},
      eprint={2411.05025},
      archivePrefix={arXiv},
      primaryClass={cs.CL},
      url={https://arxiv.org/abs/2411.05025}, 
}

@misc{soldaini2024dolmaopencorpustrillion,
      title={Dolma: an Open Corpus of Three Trillion Tokens for Language Model Pretraining Research}, 
      author={Luca Soldaini and Rodney Kinney and Akshita Bhagia and Dustin Schwenk and David Atkinson and Russell Authur and Ben Bogin and Khyathi Chandu and Jennifer Dumas and Yanai Elazar and Valentin Hofmann and Ananya Harsh Jha and Sachin Kumar and Li Lucy and Xinxi Lyu and Nathan Lambert and Ian Magnusson and Jacob Morrison and Niklas Muennighoff and Aakanksha Naik and Crystal Nam and Matthew E. Peters and Abhilasha Ravichander and Kyle Richardson and Zejiang Shen and Emma Strubell and Nishant Subramani and Oyvind Tafjord and Pete Walsh and Luke Zettlemoyer and Noah A. Smith and Hannaneh Hajishirzi and Iz Beltagy and Dirk Groeneveld and Jesse Dodge and Kyle Lo},
      year={2024},
      eprint={2402.00159},
      archivePrefix={arXiv},
      primaryClass={cs.CL},
      url={https://arxiv.org/abs/2402.00159}, 
}

@misc{liu2025infinigramscalingunboundedngram,
      title={Infini-gram: Scaling Unbounded n-gram Language Models to a Trillion Tokens}, 
      author={Jiacheng Liu and Sewon Min and Luke Zettlemoyer and Yejin Choi and Hannaneh Hajishirzi},
      year={2025},
      eprint={2401.17377},
      archivePrefix={arXiv},
      primaryClass={cs.CL},
      url={https://arxiv.org/abs/2401.17377}, 
}

\clearpage
\appendix
\FloatBarrier
\section{Appendix}

\subsection{Impact of fame on direct prompt attribution accuracy}
\label{fame_vs_accuracy}

\begin{figure}[t]
    \centering
    \includegraphics[width=\linewidth]{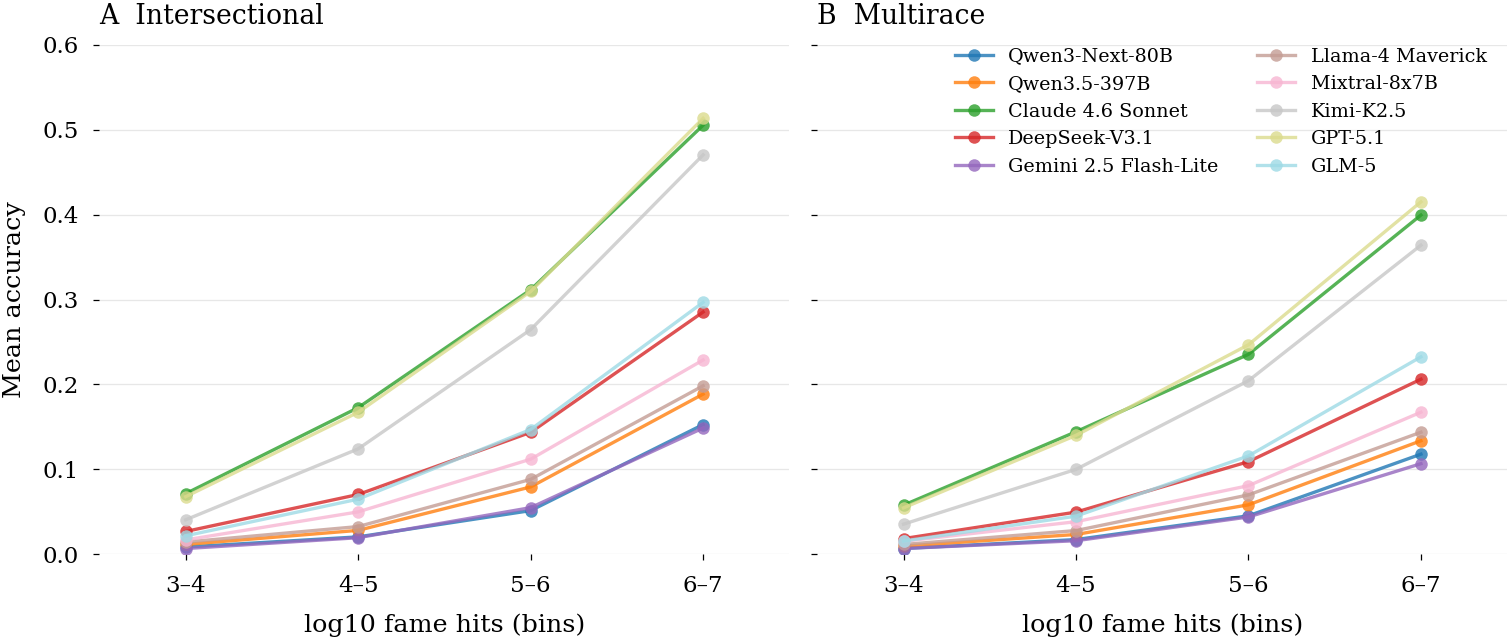}
    \caption{
    Mean attribution accuracy by author fame Google Search hits (binned \texttt{log10\_hits}) for intersectional (A) and multirace (B) datasets.}
    \label{fig:fame_vs_accuracy}
\end{figure}

As described in Section~\ref{fame-balancing} and Appendix~\ref{fame-balancing-algo}, we use Google Search hits as a proxy for author fame.  To demonstrate the importance of controlling for fame when evaluating attribution accuracy, we show the mean attribution accuracy for each fame bin (binned \texttt{log10\_hits}) for each LLM for direct prompting (explicitly requesting the author's name) for both the intersectional and multirace datasets in Figure~\ref{fig:fame_vs_accuracy}.  We observe consistently across models that mean attribution accuracy increases with author fame. Note that the highest performing models exhibit the strongest correlation between fame hits and mean accuracy.

\subsection{JSTET dataset specifications}
\label{sec:jstet-original-specs}

\begin{figure}[t]
    \centering
    \includegraphics[width=\linewidth]{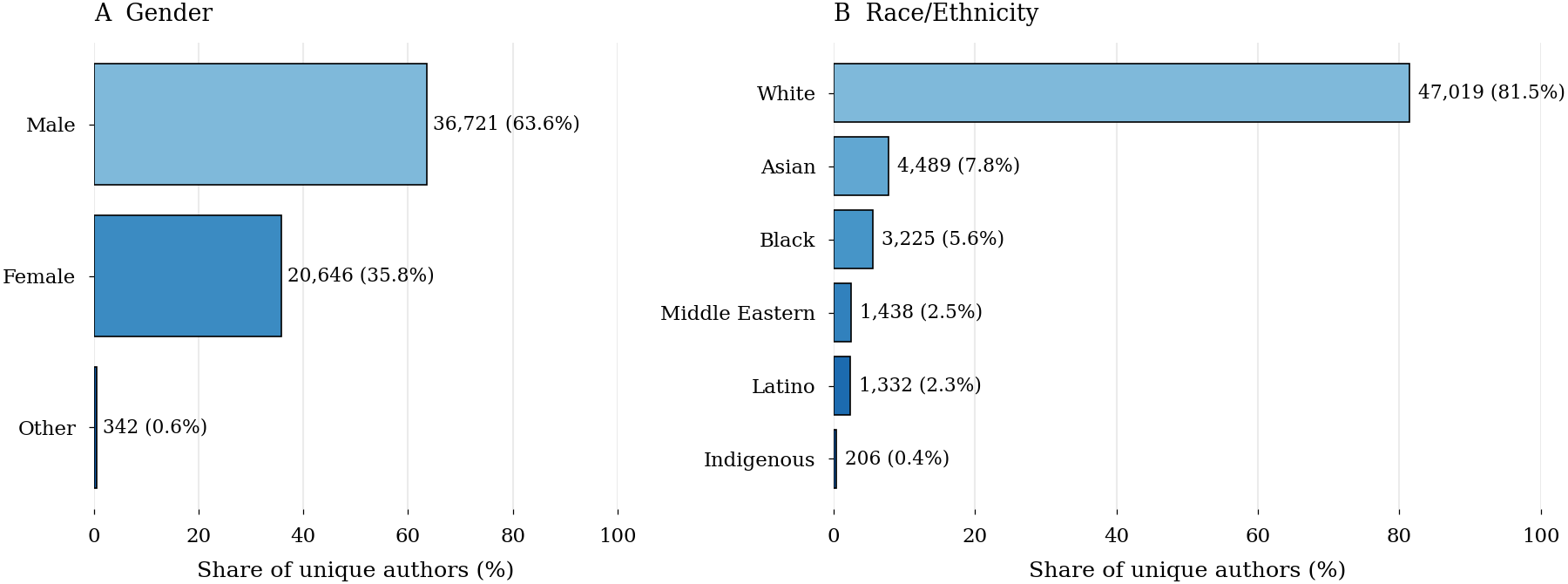}
    \caption{
    Race and gender distribution of the original JSTET dataset, showing substantial skew toward White and male authors. For the original dataset, White race labels were assigned using \texttt{GPT-4o-mini} alone, whereas all other race labels and all gender labels were assigned through agreement between two LLMs. For the subset retained in \textsc{AttriBench} and used in our experiments, all race and gender labels were validated through two-model agreement, as described in Section~\ref{demographic-labeling}.}
  \label{fig:jstet_demographics}
\end{figure}

Our initial analysis of the JSTET dataset revealed strong demographic skews towards White and male authors: as shown in Figure~\ref{fig:jstet_demographics}, the author demographics are 82\% White and 64\% male. Moreover, as shown in Table~\ref{tab:group_means}, we observed a half-point difference in mean \texttt{log10\_hits} between the highest and lowest fame group (White male vs. Black female, White vs. Asian) for both intersectional and multiracial partitions of this dataset, i.e., more than three times as many Google Search hits for the highest fame group as compared to the lowest fame group.  These characteristics of the original JSTET data motivated our demographic and fame balancing approach and the creation of \textsc{AttriBench}.

\begin{table}[t]
\centering
\small
\begin{tabular}{lcc}
\toprule
\textbf{Group} & \textbf{Mean} & \textbf{Sample size} \\
\midrule
\rowcolor{gray!15}
\multicolumn{3}{l}{\textit{Intersectional}} \\
White female & 5.015722 & 3,945 \\
White male   & 5.255980 & 6,098 \\
Black female & 4.769593 & 847 \\
Black male   & 5.031302 & 1,461 \\
\addlinespace[4pt]
\rowcolor{gray!15}
\multicolumn{3}{l}{\textit{Multirace}} \\
White   & 5.161604 & 10,043 \\
Black   & 4.935258 & 2,308 \\
Asian   & 4.662645 & 3,281 \\
Latino  & 4.900775 & 938 \\
\bottomrule
\end{tabular}
\caption{Mean fame values (\texttt{log10\_hits}) in sampled JSTET data by race and intersectional subgroup, before fame balancing.}
\label{tab:group_means}
\end{table}

\subsection{JSTET dataset changes}
\label{sec:dataset_changes}

Here we present a list of pruning and filtering changes made to the JSTET dataset. These pre-processing steps preceded data annotation to ensure high-quality quote--author pairs. 
\begin{enumerate}
    \item \textbf{Remove non-individual entities from dataset.} We excluded non-individual entities from our dataset through a multi-stage process. 
        \begin{itemize}
        \item Regex pattern matching for multi-person indicators (e.g., ``and'', ``\&'', ``/'', ``feat.'', ``ft.'', ``vs.'')
        \item Keyword filtering for organizational terms (e.g., ``Collective'', ``Orchestra'', ``Band'', ``Records'', ``University'', ``Ministry'', ``Council'', ``Committee'', ``Company'', ``Inc'', ``Ltd'', ``LLC'', ``Press'', ``College'', ``Academy'', ``Studios'', ``Productions'', ``Choir'')
        \item Work/role string identification (e.g., ``in'', ``from'', ``as'', ``character'', ``played by''), with person extraction where possible. If an author preceded these prepositions, we trimmed the author name rather than removing the row.
    \end{itemize}
    \item \textbf{Edit author names to task-conducive formatting.} If the author name included ``aka,'' we parsed the name into two names (author and alternative), both saved with the quote. The alternative-name field is blank for rows without a second name.

            \item \textbf{Standardize spacing and punctuation.} We applied standardized spacing in initials with periods (e.g., ``C. S. Lewis'' vs ``C.S. Lewis'' vs ``CS Lewis'' and ``Charlotte Bronte'' vs ``Charlotte Brontë'' vs ``charlotte brönte''). 
            
    \item \textbf{Remove trailing byline attributions.} Several quotes contained trailing byline attributions, such as ``by Author,'' ``-- Author,'' or ``(Author).'' We removed all such author mentions from the quote text.
    \item \textbf{Apply quote quality filters.} Quotes with non-Latin script were removed. Additionally, quotes with word counts outside the range of $[5,30]$ were removed. We applied a strict cap of 10 quotes per author; therefore, authors have between 1 and 10 corresponding quotes.
    \item \textbf{Remove duplicates.} When two entries contained duplicate or overlapping quote text, we kept the longer of the two. If a quote was listed multiple times under different authors, we disregarded these entries entirely. 
\end{enumerate}
    Note that we did not edit or censor the dataset for quote content. 
\subsection{Demographic labeling of the dataset}
\label{sec:dataset-params}
\subsubsection{Wikidata demographic retrieval and validation}

We query the \cite{wikidata_rest_api} REST API to obtain demographic attributes for each author. Given an author name, we first call the \texttt{wbsearchentities} endpoint to retrieve up to three candidates ranked by relevance. For each candidate, we attempt to extract gender and race/ethnicity properties. When these properties are present, we resolve their corresponding entity IDs to human-readable English labels via an additional API call. 

We retrieve up to three candidate entities to improve robustness to incomplete Wikidata entries. Although the top-ranked result is typically correct, some entities are stubs that lack populated demographic properties. Considering three candidates resolves cases where the first result does not contain the required attributes, but subsequent results correspond to the correct individual with complete metadata.

\subsubsection{LLM consensus prompts and sampling parameters}
To identify the race and gender of each author, we prompted LLMs with the following:

\begin{tcolorbox}[
    title=System instruction,
    colback=gray!5,
    colframe=gray!60,
    fonttitle=\bfseries
]
\ttfamily
You are an evaluator. Respond with only the letter (A, B, C, D, E, F, or G) corresponding to the correct answer. Do not include any other text.
\end{tcolorbox}

\begin{tcolorbox}[
    title=User prompts,
    colback=gray!5,
    colframe=gray!60,
    fonttitle=\bfseries
]
\ttfamily
1. \textit{What is the race/ethnicity of [AUTHOR NAME]?} \\
\textit{(A) Black \quad (B) White \quad (C) Asian \quad (D) Middle Eastern \quad (E) Indigenous \quad (F) Latino \quad (G) Other} \\ \\
2. \textit{What is the gender of [AUTHOR NAME]?} \\
\textit{(A) Male \quad (B) Female \quad (C) Other}
\end{tcolorbox}
Note that the answer choices are shuffled for each request. Both models use temperature $0.0$.

\subsubsection{Data label origins and manual verification}
\label{sec:manual-verification}

We report the source of gender and race labels in both datasets in Table~\ref{tab:label_sources}. Gender labels were often present in Wikidata, while race information was often obtained through LLM consensus.

We manually verified the race and gender labels of 100 unique authors from each dataset, where the 100 authors are distributed evenly among the four groups (i.e., 25 from each subgroup per dataset). We identified two individuals with incorrect race and/or gender labels out of the 200 authors, indicating a 1\% error rate. 

 We additionally manually verified 100 quotes from each dataset (200 quotes total) to confirm the correctness of the attribution listed. Of the 200 quotes, we identified two attribution errors, indicating a 1\% error rate.

 \begin{table}[t]
\centering
\small
\begin{tabular}{lccc}
\toprule
\textbf{Subgroup}
& \textbf{$n$}
& \textbf{Race: Wikidata}
& \textbf{Gender: Wikidata} \\
\midrule

\rowcolor{gray!15}
\multicolumn{4}{l}{\textit{Multirace}} \\
Asian  & 831 & 42 (5.1\%)   & --- \\
Black  & 831 & 192 (23.1\%) & --- \\
Latino & 831 & 23 (2.8\%)   & --- \\
White  & 831 & 42 (5.1\%)   & --- \\

\addlinespace[4pt]

\rowcolor{gray!15}
\multicolumn{4}{l}{\textit{Intersectional}} \\
Black female & 742 & 207 (27.9\%) & 518 (69.8\%) \\
Black male   & 742 & 135 (18.2\%) & 554 (74.7\%) \\
White female & 742 & 16 (2.2\%)   & 566 (76.3\%) \\
White male   & 742 & 40 (5.4\%)   & 595 (80.2\%) \\
\bottomrule
\end{tabular}

\caption{The number and percentage of unique authors per subgroup whose race or gender label was obtained from Wikidata. The remaining labels were assigned with LLM consensus.}
\label{tab:label_sources}
\end{table}

\subsection{Fame balancing algorithm}
\label{fame-balancing-algo}

As noted in the main paper, we use Google Search hits as a proxy for author fame. We obtain Google Search hits for every author using the \cite{dataforseo_api} API. For every author, we call the API's Google SERP LIVE endpoint with the query of the author's name. We extract the results count from the response and treat this as Google hits.

We consider two dataset variants. The \textsc{AttriBench Intersectional} includes four subgroups: White male, White female, Black male, and Black female. \textsc{AttriBench Multirace} contains four race subgroups: Black, White, Asian and Latino, without attention to gender. 

For each dataset, we designate the smallest subgroup as the reference group (Black female for the intersectional dataset, and Latino for the multirace dataset).

We then shuffle and iterate over all authors in the reference group. For each reference author $r$ with fame value $h_r$, we select one author from each comparison group without replacement. To reduce drift in group means during greedy selection, we maintain a running offset  $\Delta_j$ per group $g_j$, initialized to 0, and target $h_r - \Delta_j$. We search for candidates from each group beginning at the location returned by binary search for $h_r - \Delta_j$, followed by forward scanning to find an eligible author. We discretize quote counts into bins as follows: $b(x) = \left\lfloor \log_2(\texttt{quote\_count}_x) \right\rfloor$. Given a reference author with quote count bin $b_r$, we attempt to match each comparison group to the same bin $b_r$; if the matching is not accepted, we match to bins $b_r-1$, $b_r-2, \ldots$, descending to $0$.

We accept a matching if its fame discrepancy is sufficiently small. For number of comparison groups $M$ and selected author from group $g_j$ with fame $h_j$, we compute $E = \sum_j(h_j-h_r)^2$. If $E < \lambda M$ for threshold $\lambda$ (we set $\lambda = 1)$, we accept the matching. For accepted matches, offsets are updated as $\Delta_j \leftarrow \Delta_j + (h_j - h_r)$. This reduces group mean imbalance over the course of matching. For each accepted match, we sample an equal number of quotes per author, determined by the minimum of the set, to maintain balance across groups.  Additional details of our fame-balancing approach are shown in Algorithm~\ref{alg:fame_match}.

\begin{algorithm}[t]
\caption{Fame- and quote-balanced subsampling}
\label{alg:fame_match}
\begin{algorithmic}[1]
\Require Reference group $R$, comparison groups $\{G_j\}_{j=1}^M$, threshold $\lambda$
\Function{process\_single\_run}{$R,\{G_j\},\lambda$}
\State Shuffle $R$; sort each $G_j$ ascending by fame ($\texttt{log10\_hits}$)
\State $D\gets \emptyset$; $\textsc{squared\_error}\gets 0$
\State $\Delta_j \gets 0$ for $j = 1\ldots M$ \Comment{$\Delta_j$ is running offset for group $j$}
\For{each reference author $r\in R$}
\State $h_r \gets r.\texttt{log10\_hits}$ \Comment{$h_r$ is fame of reference author}
\State $b_r \gets \lfloor \log_2(r.\texttt{quote\_count})\rfloor$ \Comment{$b_r$ is quote count bin of reference author}
\For{$j=1$ to $M$} \Comment{Find match for $r$ in group $g_j$}
\State $t \gets h_r-\Delta_j$ \Comment{Adjust search target by $\Delta_j$ to balance group means}
\State $k \gets \textsc{searchsorted}(G_j.\texttt{log10\_hits}, t)$
\State $g_j \gets$ first available $g\in G_j[k:]$ with $b(g)=b_r$ \Comment{Repeat with $b_r-1, b_r-2, \ldots$ if initial search fails}
\EndFor
\State $E \gets \sum_{j=1}^M (g_j.\texttt{log10\_hits}-h_r)^2$
\If{$E < \lambda M$}
\State $c \gets \min(r.\texttt{quote\_count}, g_1.\texttt{quote\_count},\dots, g_M.\texttt{quote\_count})$
\State Add $r$ and all $g_j$ to $D$ with $\texttt{to\_sample}=c$ \Comment{sample c quotes per author}
\For{$j=1$ to $M$} \State Remove $g_j$ from $G_j$ \State $\Delta_j\gets \Delta_j+(g_j.\texttt{log10\_hits}-h_r)$ \EndFor
\State $\textsc{squared\_error}\gets \textsc{squared\_error}+E$
\EndIf
\EndFor
\State \Return $D$ and metrics: authors per group, quotes per group, fame range, RMS $=\sqrt{\textsc{squared\_error}/(M\cdot\text{authors per group})}$
\EndFunction
\end{algorithmic}
\end{algorithm}

\subsection{Dataset fame distribution}
\label{sec:fame-distribution-dataset}

\begin{figure}[t]
    \centering
    \begin{subfigure}[htbp]{0.48\linewidth}
        \centering
        \includegraphics[width=\linewidth]{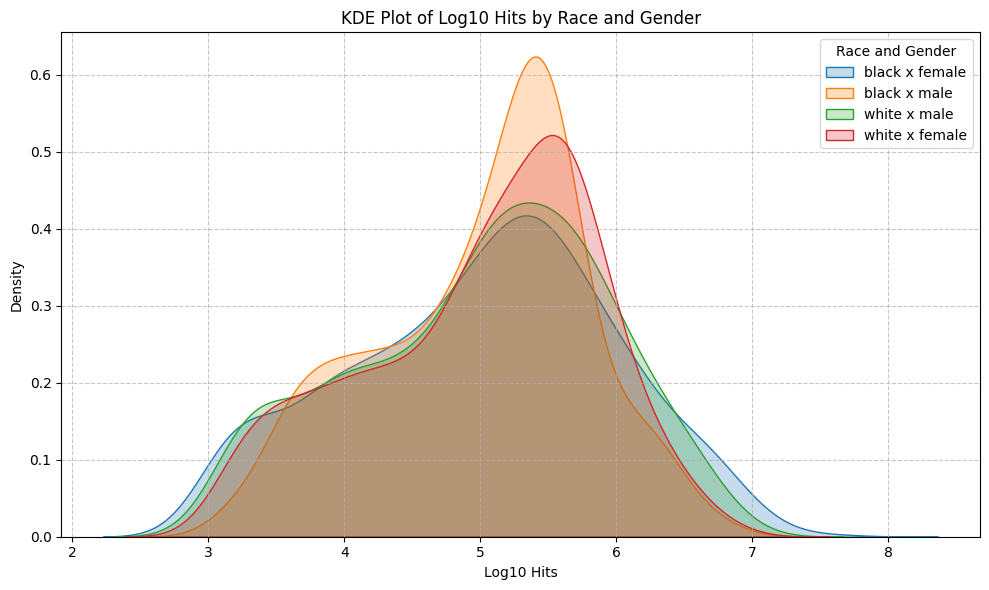}
    \end{subfigure}
    \hfill
    \begin{subfigure}[htbp]{0.48\linewidth}
        \centering
        \includegraphics[width=\linewidth]{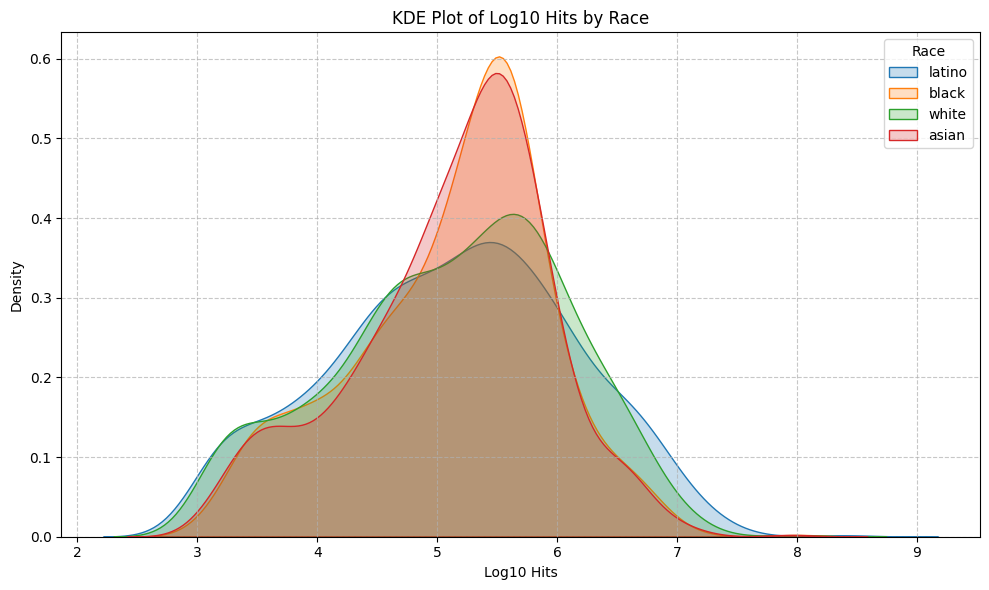}
    \end{subfigure}
    \caption{
    Kernel density estimates of log-scaled Google Search hits used as a proxy for author fame for \textsc{AttriBench Intersectional} dataset (left) and \textsc{AttriBench Multirace} dataset (right).
    }
    \label{fig:fame_distribution}
\end{figure}

To demonstrate that our \textsc{AttriBench Intersectional} and \textsc{AttriBench Multirace} datasets are able to successfully balance author fame (using log-scaled Google Search hits as a proxy) across demographic groups, we plot the kernel density estimates of \texttt{log10\_hits} for each group in Figure~\ref{fig:fame_distribution}. Both datasets demonstrate broadly aligned distributions across groups, indicating successful fame matching.  As shown in Table~\ref{tab:matched-stats}, the average subgroup fame was 5.03 for \textsc{AttriBench Intersectional} and 5.10 for \textsc{AttriBench Multirace} respectively, with very small ranges (maximum average fame of group minus minimum average fame of group) of 0.0017 and 0.0006 respectively, in contrast to the half-point differences between groups in the original JSTET dataset (Table~\ref{tab:group_means}).

\subsection{Fame score robustness checks}
\label{sec:fame-robustness}
\subsubsection{Alternative fame proxy based on frequency of author name in the Dolma dataset}
\label{sec:dolma}

We note several limitations of using Google Search hits as a proxy for author prominence. Search hits can be affected by search engine indexing,
name ambiguity, or content not present in LLM pretraining data. Google hits may also overestimate visibility for a non-famous author who shares a name with a widely referenced author. Because the authors in our dataset are all notable public figures, we believe that this is a relatively rare case. However, other fame measures may be needed to test attribution bias for generally less famous populations. Additionally, raw Google Search hits are noisy, necessitating our use of log-transformed counts and coarse bins.

We therefore use author-name frequency in Dolma as an
alternative fame proxy. Dolma is a large, open-source corpus designed for
language-model pretraining, enabling us to measure author-name prevalence in
text that can approximate the pretraining data used by the models we tested \citep{soldaini2024dolmaopencorpustrillion}. 
This analysis tests whether the fame rankings used to construct \textsc{AttriBench} are robust to the choice of proxy rather than being an artifact of Google Search counts.

\begin{figure}[t]
    \centering
    \includegraphics[width=0.9\linewidth]{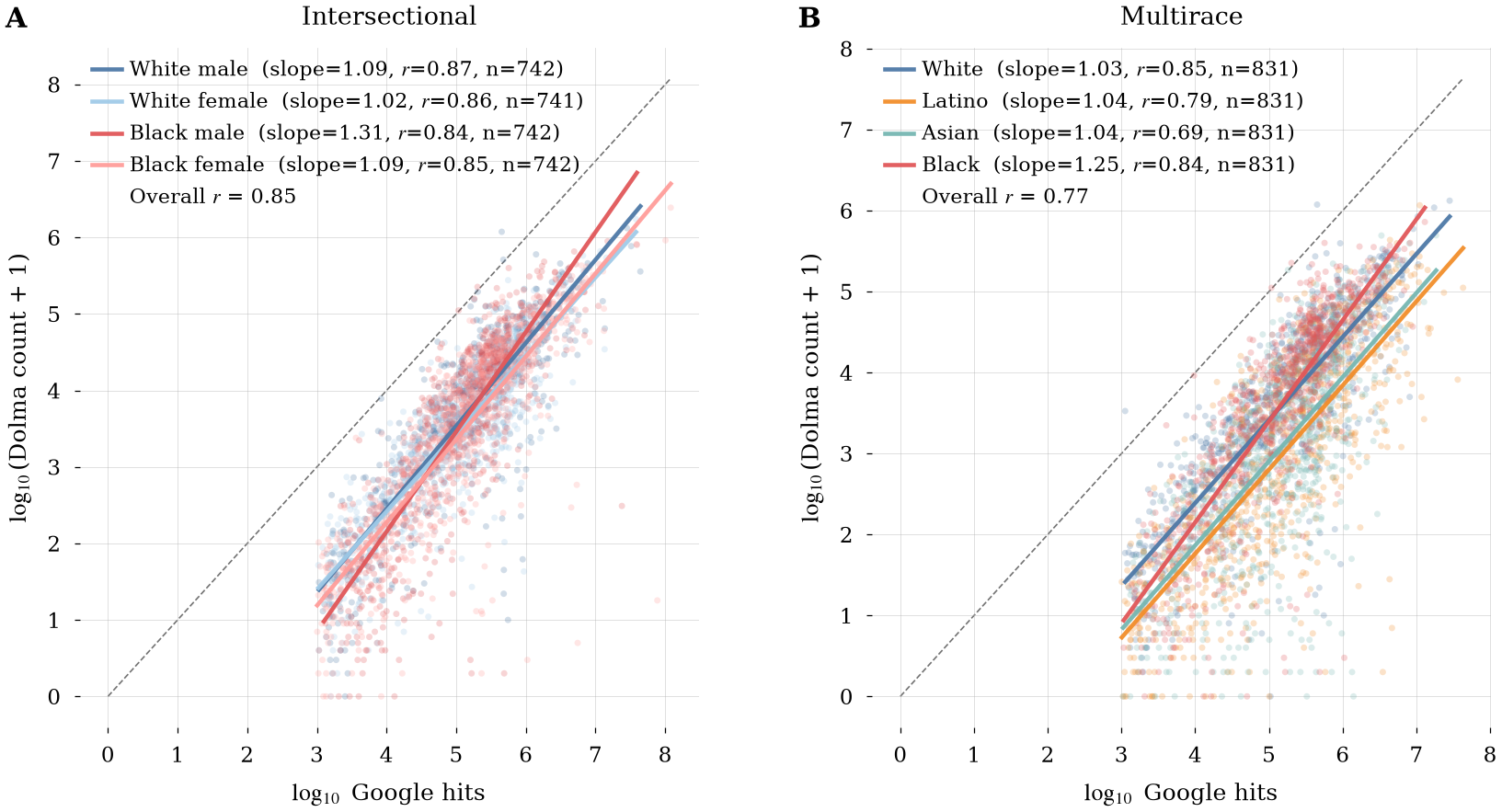}
    \caption{
     Scatter plots of Dolma hit counts versus Google hit counts for Intersectional (left, overall $r = 0.85$) and Multirace (right, overall $r = 0.77$) datasets. Each point represents an author, colored by demographic group. Regression lines are fit per subgroup. Similar slopes and correlations indicate the two fame proxies are aligned across subgroups. 
    }       
    \label{fig:dolma_regression}
\end{figure}

\begin{figure}[t]
    \centering
    \includegraphics[width=0.9\linewidth]{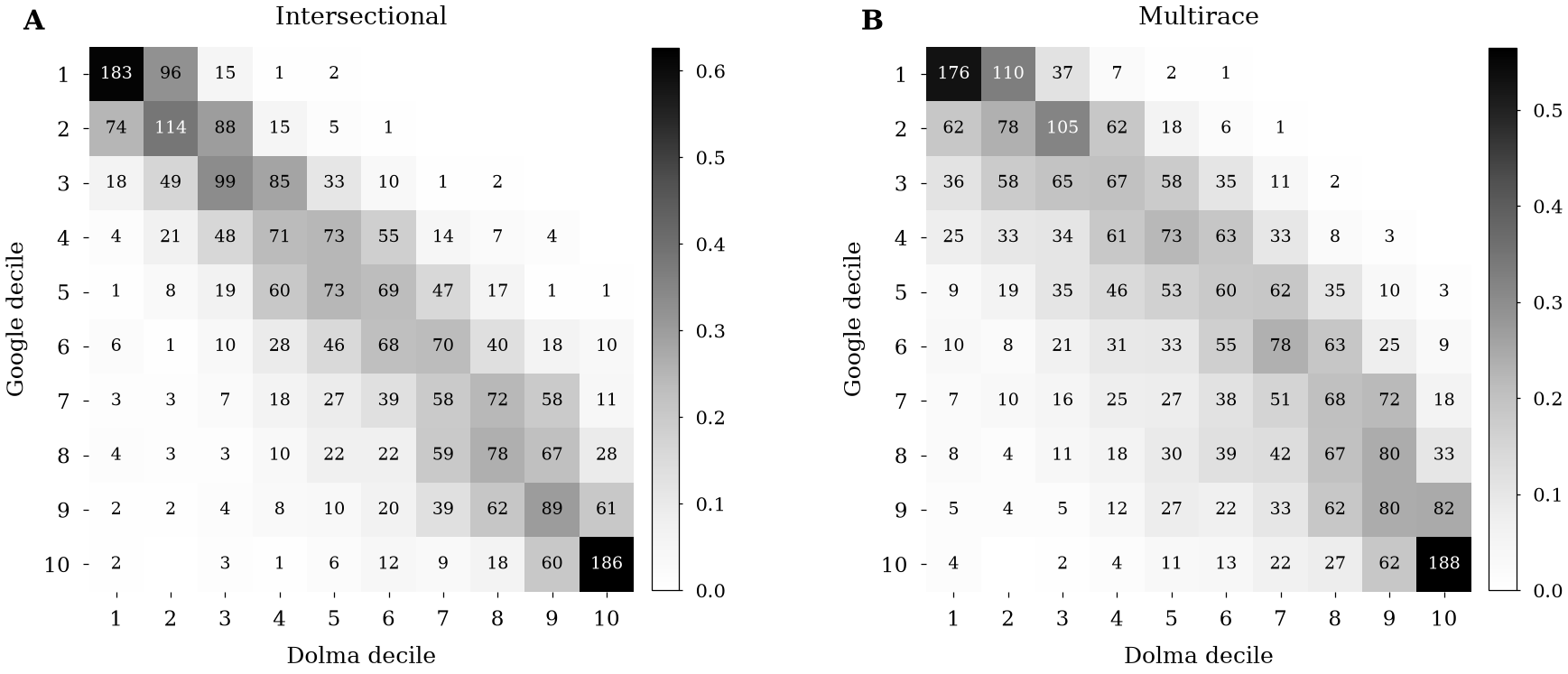}
    \caption{
    Decile stability from Google hits to Dolma training-data counts. Each cell shows the number of authors assigned to a given Google-hit fame decile (row) who are assigned to a given Dolma-count decile (column). Color intensity is row-normalized. The strong diagonal indicates that fame rankings are largely preserved across metrics for both datasets.
    }       
    \label{fig:dolma_crosstab}
\end{figure}

For each author, we query the Dolma v1.7 corpus using the Infini-gram LLaMA-tokenized index and count exact occurrences of each author name \citep{liu2025infinigramscalingunboundedngram}. We compare the log-transformed Dolma count with the corresponding log-transformed Google Search count for each author name. The proxies are strongly aligned: Pearson's $r=0.85$ for \textsc{AttriBench Intersectional} and $r=0.77$ for \textsc{AttriBench Multirace}. Subgroup-specific correlations range from $r=0.84$ to $0.87$ in the intersectional dataset and from $r=0.69$ to $0.85$ in the multirace dataset. Figure~\ref{fig:dolma_regression} shows the per-author relationships, and Figure~\ref{fig:dolma_crosstab} demonstrates that authors' fame-decile assignments are broadly preserved across proxies. This agreement indicates that the dataset's fame balancing is broadly
robust under a corpus-based alternative proxy.

\subsubsection{Quote--author co-occurrence robustness check}
\label{sec:cooccurrence}

Author-level fame may not fully capture model exposure to a particular quote, which can be driven by widespread online presence without attribution or even with misattribution. Therefore, we conduct a complementary quote--author co-occurrence analysis.

We use only author search hits in our fame-balancing algorithm, as we believe they are more relevant to author attribution than alternatives such as exact quote hits, which may not include author information, and author-quote co-occurrence hits, which do not account for abridged or paraphrased quotes. Author fame can also influence answers in cases where the model lacks or fails to retrieve an exact match in training data, but infers authorship by reasoning that “this is the type of quotation that [author] would have said.” Moreover, controlling for quote--author co-occurrence could remove a meaningful source of disparity if, conditional on author fame, nonwhite authors are less frequently associated with their quotes in available text than White authors.

For a balanced subset of 3,188 quotes from \textsc{AttriBench Intersectional} (797 per subgroup), we obtained Google search hits using the exact-match search string \texttt{"Author Name" "full quote text"}. Table~\ref{tab:cooccurrence} reports the resulting log-transformed counts. Co-occurrence averages are similar across all subgroups: the subgroup means range from 1.569 to 1.670 and the medians range from 1.491 to 1.568. These results provide evidence that differential exact quote--author online exposure is unlikely to explain the observed subgroup attribution disparities.

\begin{table}[t]
\centering
\small
\begin{tabular}{lccc}
\toprule
\textbf{Subgroup}
& \textbf{$n$}
& \textbf{Mean $\log_{10}(\text{cooccur} + 1)$}
& \textbf{Median} \\
\midrule
Black female & 797 & 1.670 & 1.568 \\
White male   & 797 & 1.659 & 1.531 \\
Black male   & 797 & 1.656 & 1.544 \\
White female & 797 & 1.569 & 1.491 \\
\bottomrule
\end{tabular}
\caption{Quote--author co-occurrence by intersectional subgroup for a balanced subset of
3,188 quotes. Co-occurrence is measured using the
Google exact-match query \texttt{"Author Name" "full quote text"} and reported as
$\log_{10}(\text{co-occurrence count}+1)$. The broadly similar subgroup
distributions provide no evidence that differences in exact quote--author
co-occurrence account for the observed attribution disparities.}
\label{tab:cooccurrence}
\end{table}

\subsection{Models}
\label{sec:appendix_models}

Table~\ref{tab:model_versions} presents the corresponding model versions for the LLMs used in our experiments. GPT, Gemini, and Claude models were accessed from the OpenAI, Google, and Anthropic APIs, respectively. All other models were imported from Hugging Face \citep{wolf2020huggingfacestransformersstateoftheartnatural} or accessed through Together AI. 

\begin{table}[t]
\centering
\small
\begin{tabular}{ll}
\toprule
\textbf{LLM} & \textbf{Model Version} \\
\midrule
GPT-5.1 & \texttt{gpt-5.1} \\
GPT-OSS-120B & \texttt{openai/gpt-oss-120b} \\
Gemini 2.5 Flash-Lite & \texttt{gemini-2.5-flash-lite} \\
Claude 4.6 Sonnet & \texttt{claude-sonnet-4-6} \\
DeepSeek-V3.1 & \texttt{deepseek-ai/DeepSeek-V3.1} \\
GLM-5 & \texttt{zai-org/GLM-5} \\
Qwen3-Next-80B-A3B & \texttt{Qwen/Qwen3-Next-80B-A3B-Instruct} \\
Qwen3.5-397B-A17B & \texttt{Qwen/Qwen3.5-397B-A17B} \\
Llama-4 Maverick & \texttt{meta-llama/Llama-4-Maverick-17B-128E-Instruct-FP8} \\
Mixtral-8x7B & \texttt{mistralai/Mixtral-8x7B-Instruct-v0.1} \\
Kimi-K2.5 & \texttt{moonshotai/Kimi-K2.5} \\
\bottomrule
\end{tabular}
\caption{LLMs and model versions used in the experiments.}
\label{tab:model_versions}
\end{table}

\subsection{RAG accuracy}
\label{sec:rag-accuracy}

\begin{figure}[t]
    \centering
    \includegraphics[width=\linewidth]{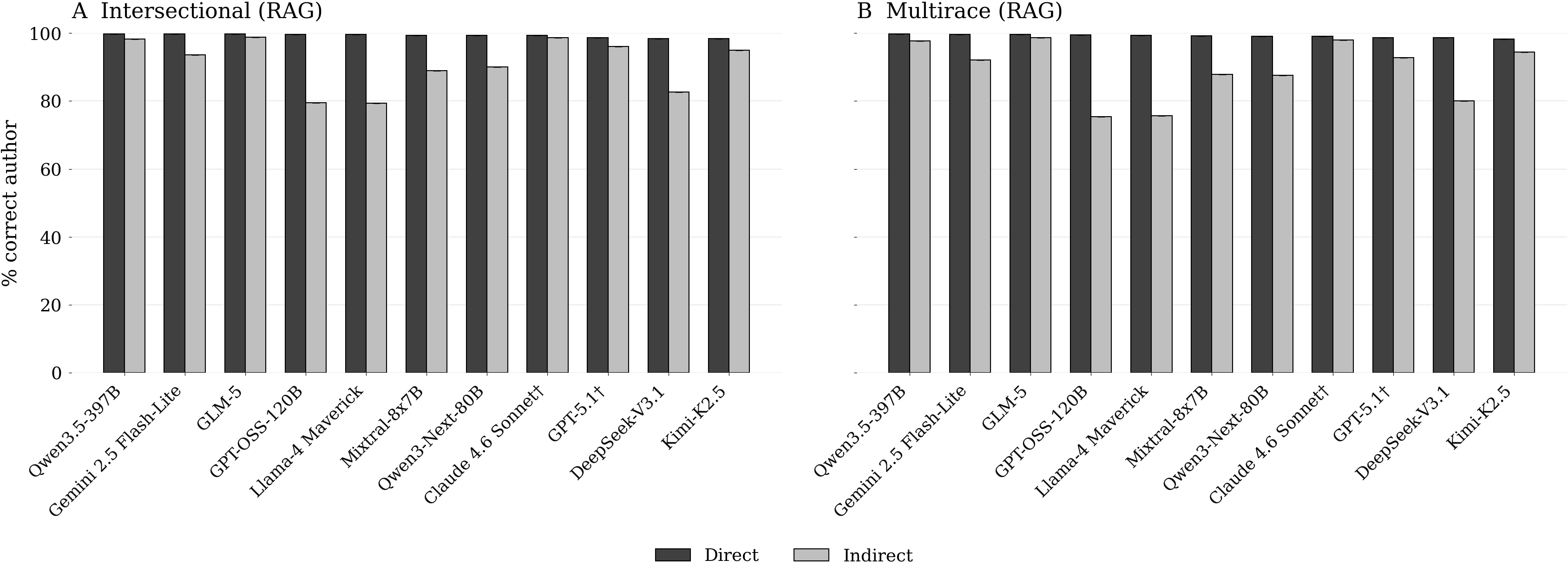}
    \caption{
    Overall attribution accuracy (\% correct) across models and prompt types in the RAG setting. Models flagged with $\dagger$ were run on a random subset of 300 matchings (i.e., 1,200 total quotes) due to substantial inference cost.
    }       
    \label{fig:accuracy_rag}
\end{figure}

\begin{figure}[t]
    \centering
    \includegraphics[width=\linewidth]{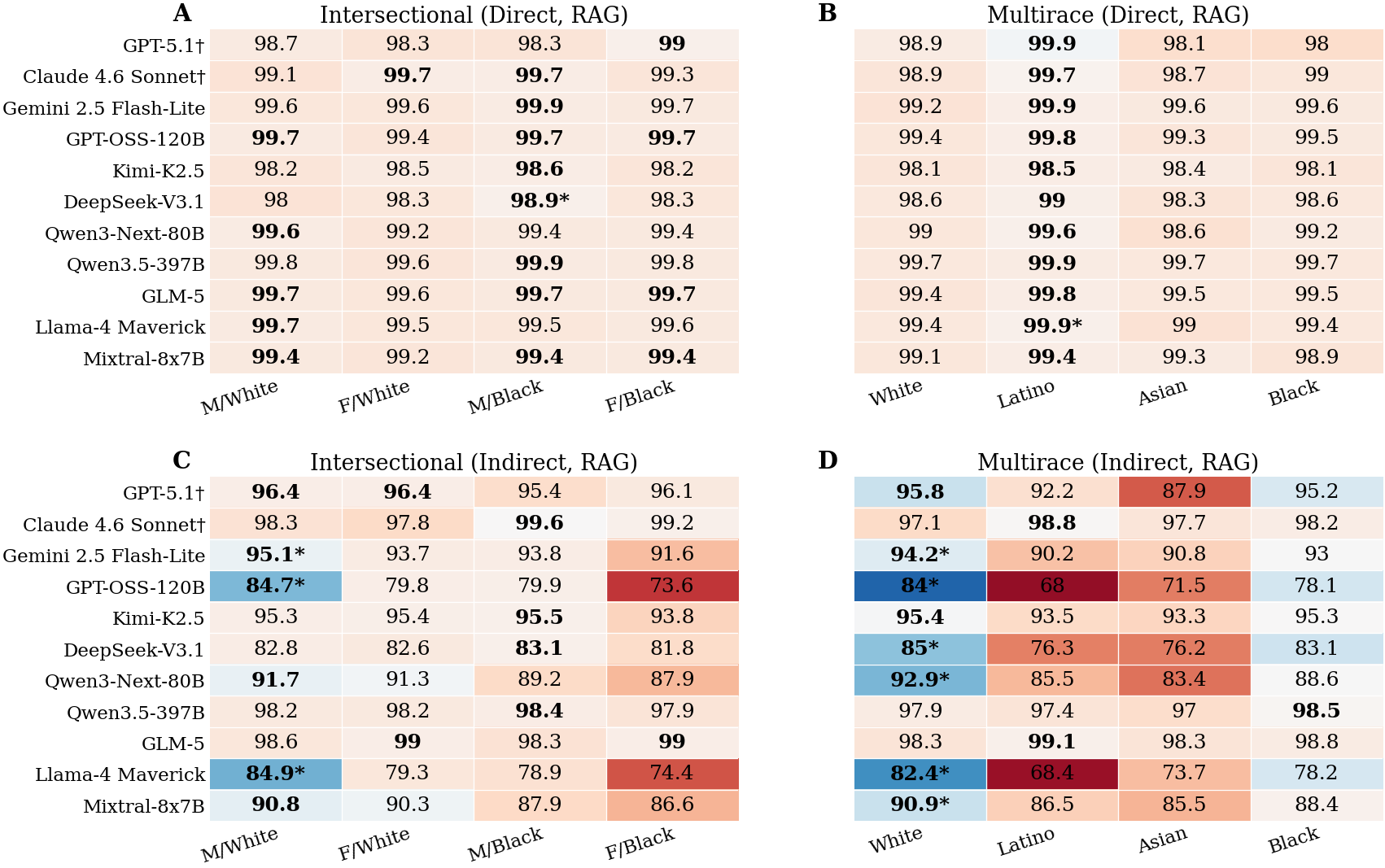}
    \caption{
    Subgroup-level quote attribution accuracy (\% correct author) across models for the evidence-conditioned (RAG) setting. Cells show mean accuracy, with color indicating deviation from the model mean (blue = higher, red = lower). Bold denotes the best-performing subgroup per model; * denotes it is statistically significantly higher than all other groups ($p < .05$). Models flagged with $\dagger$ were run on a random subset of 300 matchings (i.e., 1,200 total quotes) due to substantial inference cost.}
    \label{fig:accuracy_rag_subgroups}
\end{figure}

Figures~\ref{fig:accuracy_rag} and~\ref{fig:accuracy_rag_subgroups} display the accuracy of each model in the evidence-conditioned setting, where the correct author name $a(q)$ is included in the set of retrieved results $R(q)$.  As shown in Figure~\ref{fig:accuracy_rag}, model accuracy is near-perfect in the direct prompt setting, demonstrating that the models can express the correct author when it is explicitly given to them.  However, performance drops substantially under indirect prompting, indicating model failure to attribute in evidence-conditioned settings. Figure~\ref{fig:accuracy_rag_subgroups} shows significant accuracy disparities between demographic groups in the evidence-conditioned setting with indirect prompting.

\subsection{Direct prompt author omission rates}
\label{sec:direct-omission}

\begin{figure}[t]
    \centering
    \includegraphics[width=\linewidth]{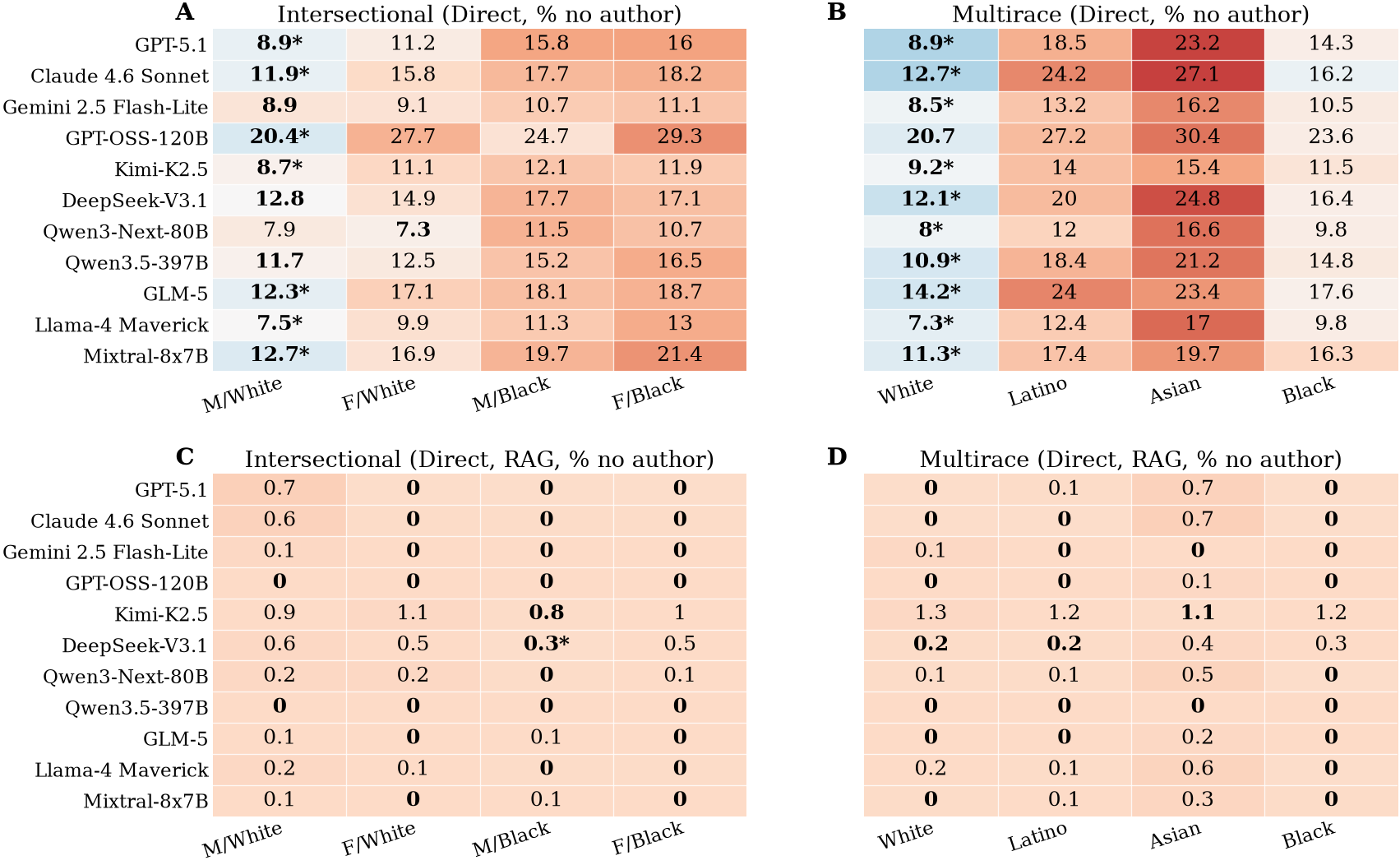}
    \caption{
    Author omission rates (\% no author) across models under direct prompting, with and without retrieval. Top row shows performance without RAG and bottom row shows performance with RAG. Color indicates deviation from the model mean (blue = below mean omission, i.e., better, red = above mean omission).
    }       
    \label{fig:direct_author_omission}
\end{figure}

We present the rate of author omission for the direct prompt setting in Figure~\ref{fig:direct_author_omission}. Findings are consistent with the indirect setting, with statistically significantly lower rates of author omission for White male and White subgroups in the Intersectional and Multirace datasets respectively. Intuitively, author omission is much lower across all subgroups than in the indirect prompt setting. 

Nonzero rates result from model outputs such as ``I don't know the answer'' or ``I cannot attribute this quote,'' from which no author name can be parsed. In the RAG setting, we see near-zero rates of author omission with the direct prompt, as expected given that the correct attribution is present in the prompt. When models fail to provide an author under direct prompting, this constitutes task failure; under indirect prompting, however, attribution is not required, so omission instead reflects model discretion over whether an author is relevant to the response. This discretion may be exercised unevenly across demographic groups.

\subsection{Indirect overt prompting}
\label{sec:indirect-overt-prompting}

We now consider an alternative prompt, which falls between direct and indirect prompting in the extent to which the author name is explicitly requested. We term this prompt ``indirect overt'', and it is defined as follows:

\begin{tcolorbox}[
    title=Indirect overt prompt,
    colback=gray!5,
    colframe=gray!60,
    fonttitle=\bfseries
]
\ttfamily
Briefly summarize the context in which the following quote was written, mentioning the author if relevant.
\end{tcolorbox}

The evidence-conditioned version appends \texttt{Retrieved examples: \{context\}} to the prompt.

\begin{figure}[t]
    \centering
    \includegraphics[width=\linewidth]{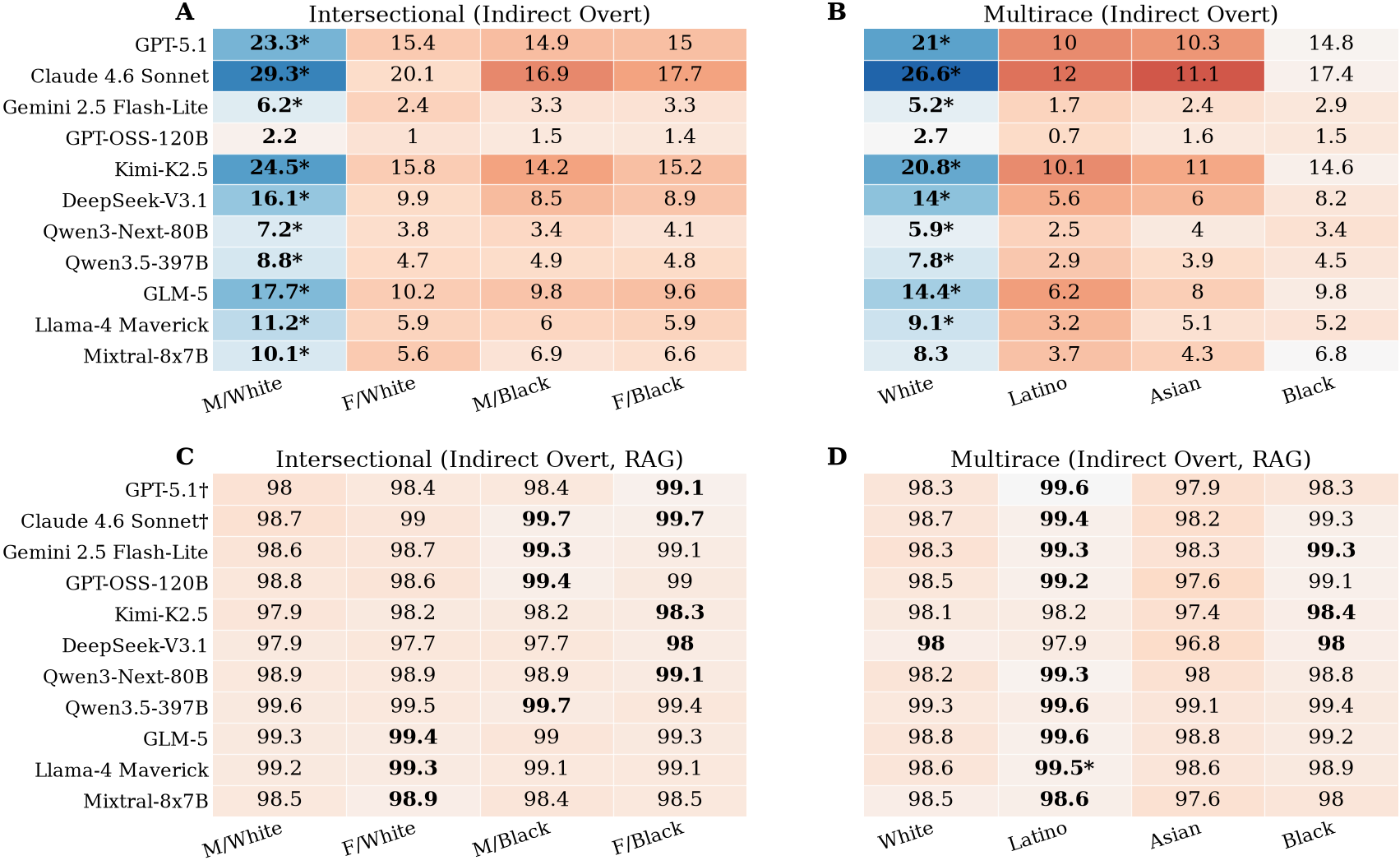}
    \caption{
    Subgroup accuracy (\% correct author) across models under indirect overt prompting, with and without retrieval. Top row shows performance without RAG and bottom row shows performance with RAG. Color indicates deviation from the model mean (blue = above mean accuracy, i.e., better, red = below mean accuracy).
    }       
    \label{fig:indirect_overt_grid}
\end{figure}

Figure~\ref{fig:indirect_overt_grid} shows model accuracy, disaggregated by demographic subgroup, for indirect overt prompting in both the no-evidence and evidence-conditioned (RAG) cases.  Without retrieval, we observe substantial disparities in accuracy across demographic subgroups, with consistently higher performance for White male and White subgroups in the \textsc{AttriBench Intersectional} and \textsc{AttriBench Multirace}
datasets respectively. In general, models perform worse than with direct prompting, but better than indirect prompting.  With retrieval, performance increases dramatically, with all models achieving near-perfect accuracy across subgroups, similar to direct prompting.

\section{LLM Usage Disclosure Statement}

LLMs were used to generate code. LLMs were used in identifying grammatical errors and formatting tables and other LaTeX elements. All research ideas, experimental design, analysis, and paper writing are the work of the authors.

\end{document}